\documentclass{article}

\usepackage{microtype}
\usepackage{graphicx}
\usepackage{booktabs} 
\usepackage{amsfonts} 
\usepackage{amsmath}
\DeclareMathOperator{\si}{\text{sim}}

\usepackage{caption}
\usepackage{subcaption}
\usepackage{adjustbox}
\usepackage{wrapfig}
\usepackage{float}

\usepackage{footnote}
\usepackage{hyperref}
\usepackage[numbers]{natbib} 

\newtheorem{definition}{Definition}[section]

\usepackage[accepted]{icml2022}

\icmltitlerunning{Geometric Multimodal Contrastive Representation Learning}

\begin{document}

\twocolumn[
\icmltitle{Geometric Multimodal Contrastive Representation Learning}



\icmlsetsymbol{equal}{*}

\begin{icmlauthorlist}
\icmlauthor{Petra Poklukar}{equal,kth}
\icmlauthor{Miguel Vasco}{equal,lisbon}
\icmlauthor{Hang Yin}{kth}
\icmlauthor{Francisco S. Melo}{lisbon}
\icmlauthor{Ana Paiva}{lisbon}
\icmlauthor{Danica Kragic}{kth}
\end{icmlauthorlist}

\icmlaffiliation{kth}{KTH Royal Institute of Technology, Stockholm, Sweden}
\icmlaffiliation{lisbon}{INESC-ID \& Instituto Superior Técnico, University of Lisbon, Portugal}

\icmlcorrespondingauthor{Miguel Vasco}{miguel.vasco@tecnico.ulisboa.pt}
\icmlcorrespondingauthor{Petra Poklukar}{poklukar@kth.se}

\icmlkeywords{Multimodal, Representation Learning, Contrastive Learning}

\vskip 0.3in
]



\printAffiliationsAndNotice{\icmlEqualContribution} 

\begin{abstract}
Learning representations of multimodal data that are both informative and robust to missing modalities at test time remains a challenging problem due to the inherent heterogeneity of data obtained from different channels. To address it, we present a novel Geometric Multimodal Contrastive (GMC) representation learning method consisting of two main components: \textit{i)} a two-level architecture consisting of \emph{modality-specific} base encoders, allowing to process an arbitrary number of modalities to an intermediate representation of fixed dimensionality, and a \emph{shared} projection head, mapping the intermediate representations to a latent representation space; \textit{ii)} a multimodal contrastive loss function that encourages the geometric alignment of the learned representations. We experimentally demonstrate that GMC representations are semantically rich and achieve state-of-the-art performance with missing modality information on three different learning problems including prediction and reinforcement learning tasks.
\end{abstract}

\section{Introduction} \label{sec:introduction}

Information regarding objects or environments in the world can be recorded in the form of signals of different nature. These different \emph{modality} signals can be for instance images, videos, sounds or text, and represent the same underlying phenomena. Naturally, the performance of machine learning models can be enhanced by leveraging the redundant and complementary information provided by multiple modalities~\cite{baltruvsaitis2018multimodal}. In particular, exploiting such multimodal information has been shown to be successful in tasks such as classification~\cite{tsai2018learning,tsai2019MULT}, generation~\cite{wu2018multimodal,shi2019variational} and control~\cite{silva2019playing, vasco2021sense}.

\begin{figure}[t!]
      \centering  
      \includegraphics[width=0.6\linewidth]{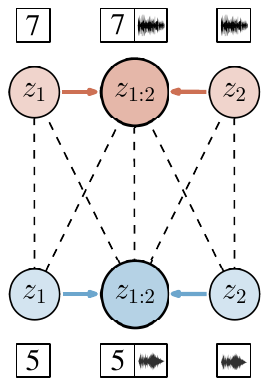}
      \caption{We propose the \emph{Geometric Multimodal Contrastive} (GMC) framework to learn representations of multimodal data by \emph{aligning} the corresponding modality-specific ($z_1$ or $z_2$) and complete ($z_{1:2}$) representations (solid arrows, blue circles) and \emph{contrasting} with different modality-specific and complete pairs (dashed lines, red circles).}
    \label{fig:gmc:intro}
\end{figure}

The advances of many of these methods can be attributed to the efficient learning of multimodal data representations, which reduces the inherent complexity of raw multimodal data and enables the extraction of the underlying semantic correlations among the different modalities~\cite{baltruvsaitis2018multimodal,multimodal-repr-learning-survey}. Generally, good representations of multimodal data \emph{i) capture the semantics} from individual modalities necessary for performing a given downstream task. 
Additionally, in scenarios such as real-world classification and control, it is essential that the obtained representations are \emph{ii) robust to missing modality} information during execution \cite{meo2021multimodal,tremblay2021multimodal,zambelli2020multimodal}. 
In order to fulfill \emph{i)} and \emph{ii)}, the unique characteristics of each modality need to be processed accordingly and efficiently combined, which remains a challenging problem known as the \emph{heterogeneity gap} in multimodal representation learning~\citep{multimodal-repr-learning-survey}.

\begin{figure*}[t]
    \centering
    \begin{subfigure}[b]{0.31\textwidth}
        \centering
        \includegraphics[height=2.8cm]{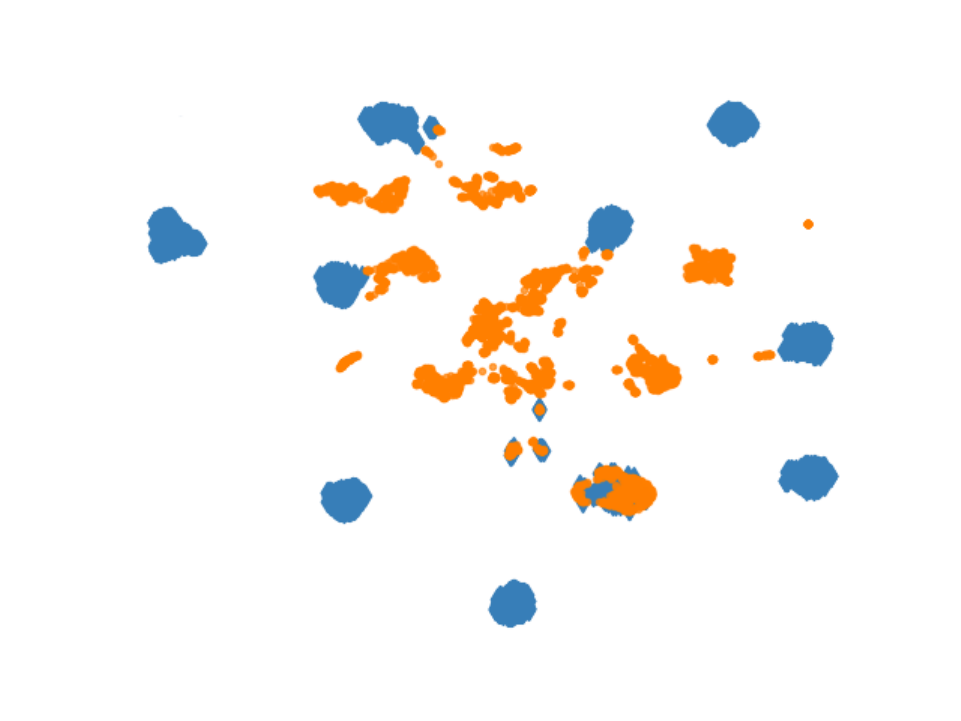}
        \caption{MVAE \cite{wu2018multimodal}}
        \label{fig:problem:mvae}
    \end{subfigure}
    \hfill
    \begin{subfigure}[b]{0.31\textwidth}
        \centering
        \includegraphics[height=2.8cm]{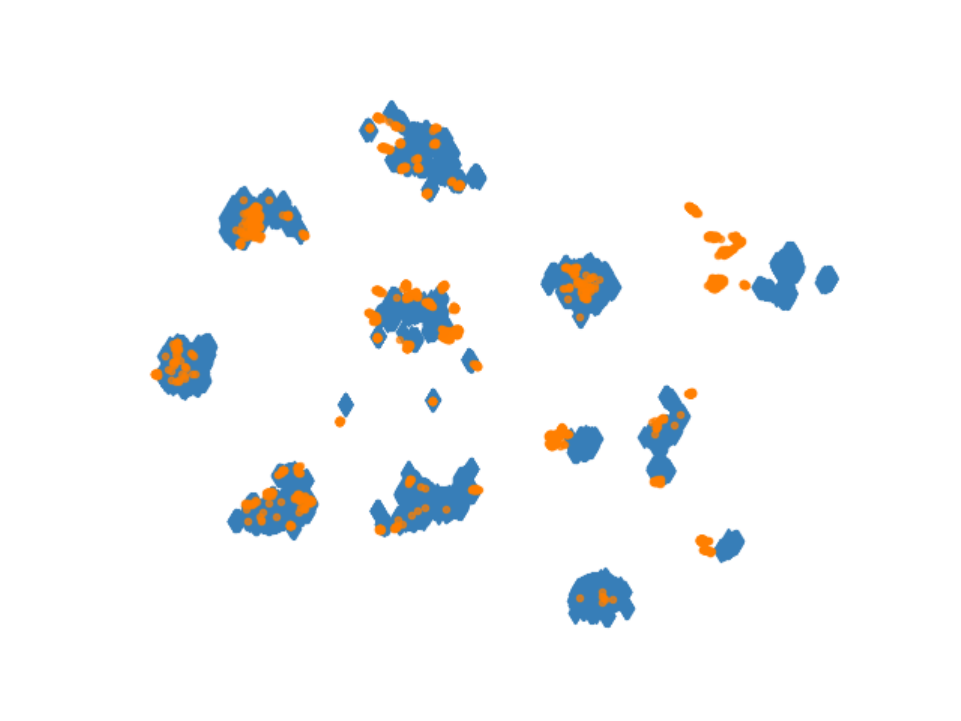}
        \caption{MMVAE \cite{shi2019variational}}
        \label{fig:problem:mmvae}
    \end{subfigure}
    \hfill
    \begin{subfigure}[b]{0.31\textwidth}
        \centering
        \includegraphics[height=2.8cm]{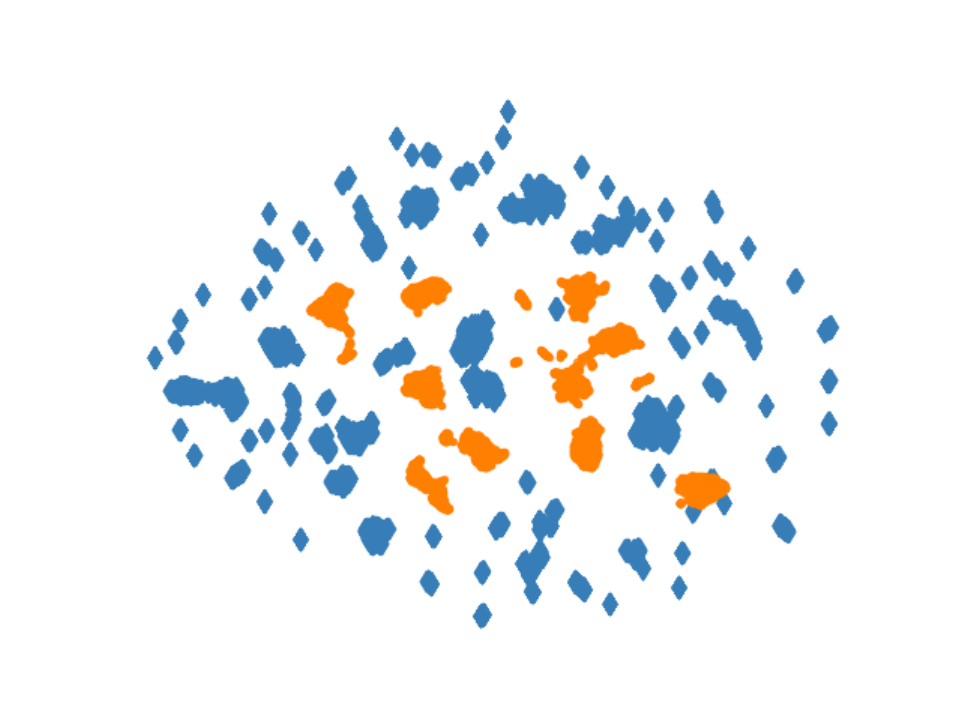}
        \caption{Nexus \cite{vasco2022leveraging}}
        \label{fig:problem:nexus}
    \end{subfigure}
    
    \vspace{1ex}
    
    \begin{subfigure}[b]{0.31\textwidth}
        \centering
        \includegraphics[height=2.8cm]{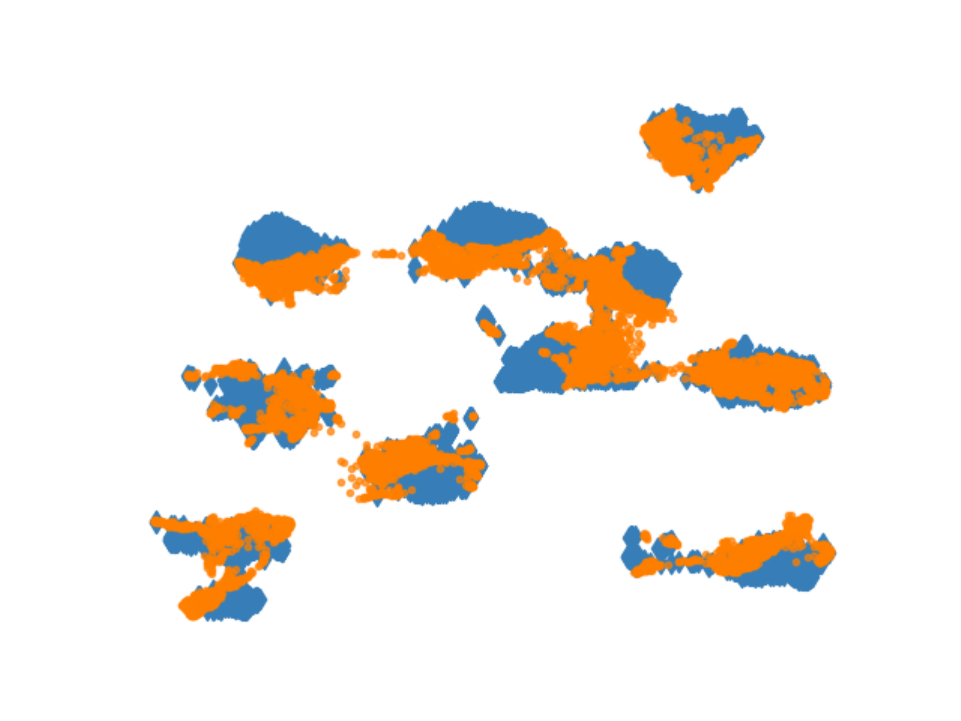}
        \caption{MUSE \cite{vasco2021sense}}
        \label{fig:problem:muse}
    \end{subfigure}
    \hfill
    \begin{subfigure}[b]{0.31\textwidth}
        \centering
        \includegraphics[height=2.8cm]{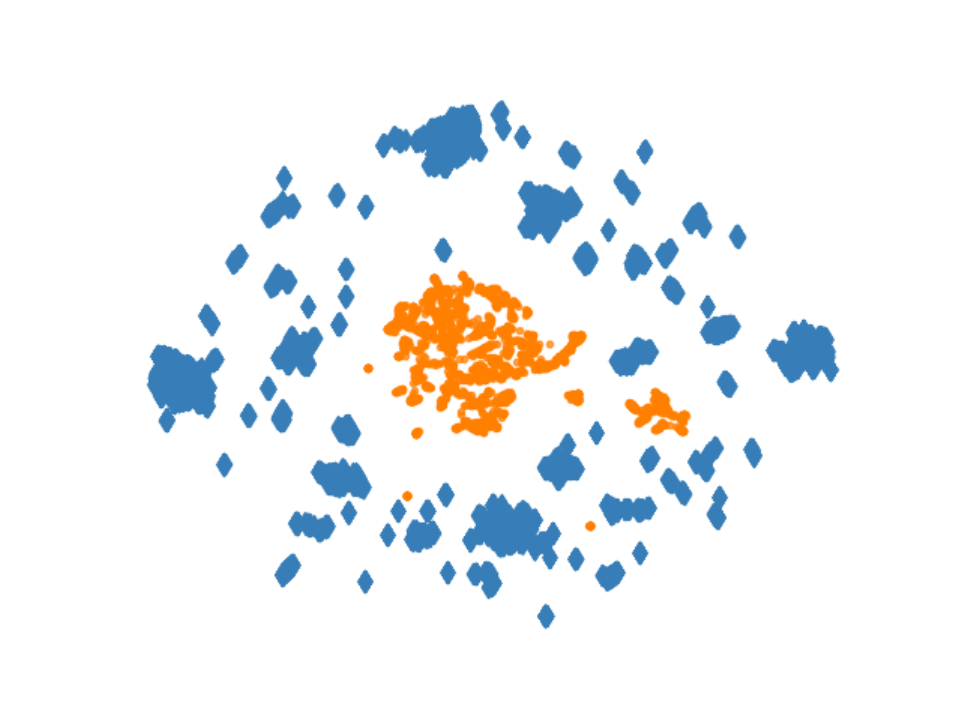}
        \caption{MFM \cite{tsai2018learning}}
        \label{fig:problem:mfm}
    \end{subfigure}
    \hfill
    \begin{subfigure}[b]{0.31\textwidth}
        \centering
        \includegraphics[height=2.8cm]{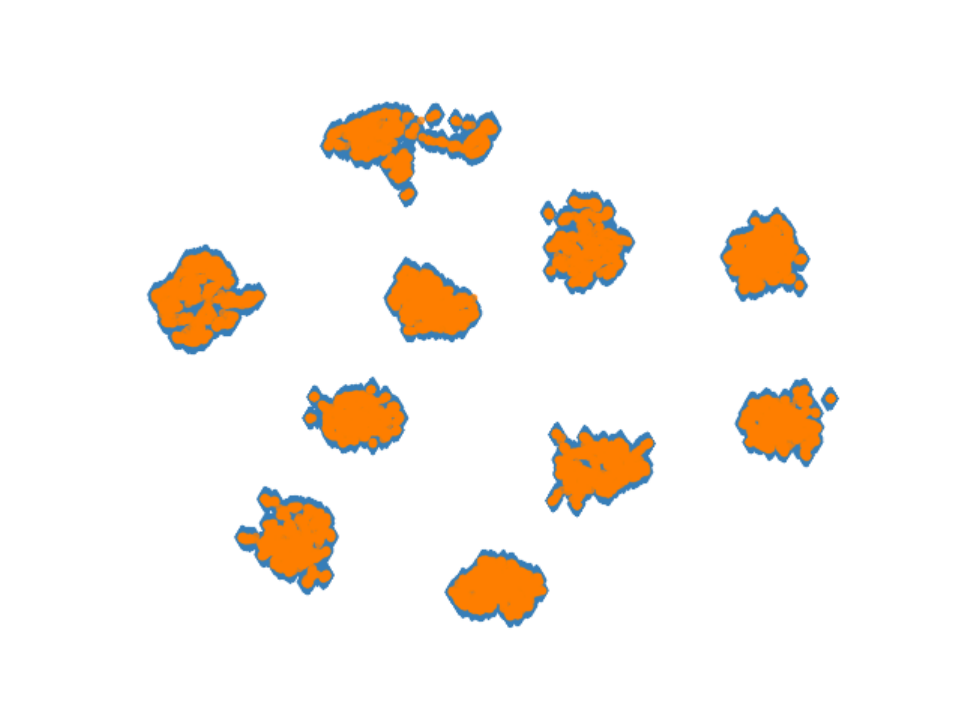}
        \caption{\textbf{GMC (Ours)}}
        \label{fig:problem:gmc}
    \end{subfigure}
    \caption{UMAP visualization of complete representations $z_{1:4}$ (blue) and image representations $z_1$ (orange) in a latent space $z \in \mathbb{R}^{64}$ obtained from several state-of-the-art multimodal representation learning models on the MHD dataset considered in Section~\ref{sec:experiments:unsupervised}. Only GMC is able to learn \emph{modality-specific} and \emph{complete} representations that are geometrically aligned. More visualizations in Appendix \ref{sec:appendix:vis}.}
    \label{fig:misalignment}
\end{figure*}

An intuitive idea to mitigate the heterogeneity gap is to project heterogeneous data into a shared representation space such that the representations of complete observations capture the semantic content shared across all modalities. In this regard, two directions have shown promise, namely, generation-based methods commonly extending the Variational Autoencoder (VAE) framework~\citep{kingma2013auto} to multimodal data such as MVAE~\cite{wu2018multimodal} and MMVAE~\cite{shi2019variational}, as well as methods relying on the fusion of modality-specific representations such as MFM~\cite{tsai2018learning} and the Multimodal Transformer~\cite{tsai2019MULT}. Fusion based methods by construction fulfill objective \textit{i)} but typically do not provide a mechanism to cope with missing modalities. While this is better accounted for in the generation based methods, these approaches often struggle to align complete and modality-specific representations due to the demanding reconstruction objective. We thoroughly discuss the geometric misalignment of these methods in Section \ref{sec:motivation}.

In this work, we learn \emph{geometrically aligned} multimodal data representations that provide robust performance in downstream tasks under missing modalities at test time. 
To this end, we present the \textit{Geometric Multimodal Contrastive (GMC)} representation learning framework. 
Inspired by the recently proposed Normalized Temperature-scaled Cross Entropy (NT-XEnt) loss in visual contrastive representation learning~\citep{simclr}, we contribute a novel multimodal contrastive loss that explicitly aligns modality-specific representations with the representations obtained from the corresponding complete observation, as depicted in Figure~\ref{fig:gmc:intro}. 
GMC assumes a two-level neural-network model architecture consisting of a collection of \emph{modality-specific} base encoders, processing modality data into an intermediate representation of a fixed dimensionality,
and a \emph{shared} projection head, mapping the intermediate representations into a latent representation space where the contrastive learning objective is applied. It can be scaled to an arbitrary number of modalities, and provides semantically rich representations that are 
robust to missing modality information. Furthermore, as shown in our experiments, GMC is general as it can be integrated into existing models and applied to a variety of challenging problems, such as learning representations in an unsupervised manner (Section \ref{sec:experiments:unsupervised}), for prediction tasks using a weak supervision signal (Section \ref{sec:experiments:supervised}) or downstream reinforcement learning tasks (Section \ref{sec:experiments:rl}). We show that GMC is able to achieve state-of-the-art performance with missing modality information compared to existing models.

\vspace{0.3cm}
\section{The Problem of Geometric Misalignment in Multimodal Representation Learning}
\label{sec:motivation}

\begin{figure*}[t]
    \centering
    \includegraphics[height=5.8cm]{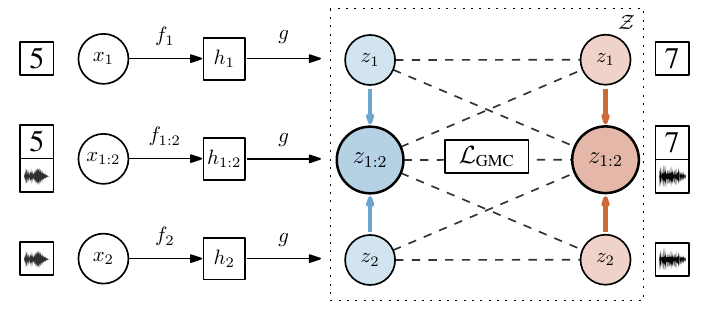}
    \vspace{-2ex}
    \caption{The \emph{Geometric Multimodal Contrastive} (GMC) framework instantiated in scenarios with two modalities ($M=2$): \emph{modality-specific} base networks $f(\cdot) = \{f_{1:2}(\cdot)\} \cup \{ f_1(\cdot), f_2(\cdot)\}$ encode common-dimensionality intermediate representations $h$ that are projected using a \emph{shared} projection head $g(\cdot)$ to a common representation space $\mathcal{Z}$, in which we apply a novel multimodal contrastive loss $\mathcal{L}_{\text{GMC}}$, detailed in Eq. \eqref{eq:contrastive}, that \emph{aligns} corresponding modality-specific $\{z_1, z_2\}$ and complete $z_{1:2}$ representations (coloured arrows) and \emph{contrasts} with representations from different observations (dashed lines).}
    \label{fig:gmc:method}
\end{figure*}

We consider scenarios where information is provided in the form of a dataset $X$ of $N$ tuples, i.e., ${X = \{x_{1:M}^i = (x_1^i, \dots, x_M^i)\}_{i = 1}^N}$, where each tuple $x_{1:M} = (x_1, \dots, x_M)$ represents observations provided by $M$ different modalities. We refer to the tuples $x_{1:M}$ consisting of all $M$ modalities as \textit{complete} observations and to the single observations $x_m$ as \textit{modality-specific}.
The goal is to learn complete representations $z_{1:M}$ of $x_{1:M}$ and modality-specific representations $\{z_1, \dots, z_M\}$ of $\{x_1, \dots, x_M\}$ that are:

\begin{itemize}
    \item[\textit{i)}] informative, i.e., both $z_{1:M}$ and any of ${z_m \in \{z_1, \dots, z_M\}}$ contains relevant semantic information for some downstream task, and thus,
    
    \item[\textit{ii)}] robust to missing modalities during test time, i.e.,  the success of a subsequent downstream task is independent of whether the provided input is the complete representation $z_{1:M}$ or any of the modality-specific representations $z_m \in \{z_1, \dots, z_M\}$.
\end{itemize}

Prior work has demonstrated success in using complete representations $z_{1:M}$ in a diverse set of applications, such as image generation \cite{wu2018multimodal,shi2019variational} and control of Atari games \cite{silva2019playing,vasco2021sense}.  
Intuitively, if complete representations $z_{1:M}$ are sufficient to perform a downstream task then learning modality-specific representations that are geometrically aligned with $z_{1:M}$ in the same representation space should ensure that $z_m$ contain necessary information to perform the task even when $z_{1:M}$ cannot be provided. 
Therefore, in Section \ref{sec:experiments}  we study the geometric alignment of $z_{1:M}$ and each $z_m$ on several multimodal datasets and state-of-the-art multimodal representation learning models. In Figure~\ref{fig:misalignment}, we visualize an example of encodings of $z_{1:M}$ (in blue) and $z_{m}$ corresponding to the image modality (in orange) where we see that the existing approaches produce geometrically misaligned representations. As we empirically show in Section \ref{sec:experiments}, this misalignment is consistent across different learning scenarios and datasets, and can lead to a poor performance on downstream tasks.

To fulfill \textit{i)} and \textit{ii)}, we propose a novel approach that builds upon the simple idea of geometrically aligning modality-specific representations $z_m$ with the corresponding complete representations $z_{1:M}$ in a latent representation space, framing it as a contrastive learning problem.

\section{Geometric Multimodal Contrastive Learning}
 
\looseness=-1
We present the \emph{Geometric Multimodal Contrastive} (GMC) framework, visualized in Figure \ref{fig:gmc:method}, consisting of three main components:
\begin{itemize}
    \item A collection of neural network \emph{base} encoders $f(\cdot) = \{f_{1:M}(\cdot)\} \cup \{ f_1(\cdot), \dots, f_M(\cdot)\}$, where $f_{1:M}(\cdot)$ and $f_m(\cdot)$ take as input the complete $x_{1:M}$ and modality-specific observations $x_m$, respectively, and output intermediate $d$-dimensional representations $\{h_{1:M}, h_1, \dots, h_M\} \in \mathbb{R}^d$;
    
    \item A neural network \emph{shared} projection head $g(\cdot)$ that maps the intermediate representations given by the base encoders $f(\cdot)$ to the latent representations $\{z_{1:M}, z_1, \dots, z_M\} \in \mathbb{R}^s$ over which we apply the contrastive term. The projection head $g(\cdot)$ enables to encode the intermediate representations in a shared representation space $\mathcal{Z}$ while preserving modality-specific semantics;
    
    \item A multimodal contrastive NT-Xent loss function ($\mathcal{L}_{\text{GMC}}$), that is inspired by the recently proposed SimCLR framework~\cite{simclr} and encourages the geometric alignment of $z_m$ and $z_{1:M}$.
\end{itemize}

We phrase the problem of geometrically aligning $z_m$ with $z_{1:M}$ as a contrastive prediction task where the goal is to identify $z_m$ and its corresponding complete representation $z_{1:M}$ in a given mini-batch. 
Let ${\mathcal{B} = \{z_{1:M}^i\}_{i = 1}^B \subset g(f(X))}$ be a mini-batch of $B$ complete representations. Let $\si(u, v)$ denote the cosine similarity among vectors $u$ and $v$ and let $\tau \in (0, \infty)$ be the temperature hyperparameter. We denote by

\begin{align}
    s_{p, q} (i, j) = \exp(\si(z_{p}^i, z_{q}^j)/\tau),
    \label{eq:similarity}
\end{align}
the similarity between representations $z_{p}^i$ and $z_{q}^j$ (modality-specific or complete) corresponding to the $i$th and $j$th samples from the mini-batch $\mathcal{B}$. For a given modality $m$, we define positive pairs as $(z_{m}^i, z_{1:M}^i)$ and $(z_{1:M}^i, z_{m}^i)$ for $i = 1, \dots, B$ and treat the remaining pairs as negative ones. In particular, we denote by
\begin{align*}
    \Omega_{p, q}(i) = \sum_{i \neq j} s_{p, p} (i, j) + \sum_{j} s_{p, q} (i, j) 
\end{align*}
the sum of similarities among negative pairs that correspond to the positive pair $(z_{p}^i, z_{q}^i)$. We define the contrastive loss for the positive pairs $(z_{m}^i, z_{1:M}^i)$ and $(z_{1:M}^i, z_{m}^i)$ as the sum
\begin{align*}
    l_m (i) = - \log \frac{s_{m, 1:M} (i, i)}{\Omega_{m, 1:M}(i)}  - \log \frac{s_{1:M, m} (i, i)}{\Omega_{1:M, m}(i)}.
\end{align*}
Lastly, we combine the loss terms for each modality ${m = 1, \dots, M}$ and obtain the final training loss
\begin{align}
    \mathcal{L}_{\text{GMC}}(\mathcal{B}) = \sum_{m = 1}^M \sum_{i = 1}^B l_m(i).
    \label{eq:contrastive}
\end{align}

As we only contrast single modality-specific representations to the complete ones, $\mathcal{L}_{\text{GMC}}$ scales linearly to an arbitrary number of modalities. In Section \ref{sec:experiments}, we show that $\mathcal{L}_{\text{GMC}}$ can be added as an additional term to existing frameworks to improve their robustness to missing modalities. Moreover, we experimentally demonstrate that the architectures of the base encoders and shared projection head can be flexibly adjusted depending on the task.
\section{Related Work}
\label{section:related_work}
Learning multimodal representations suitable for downstream tasks has been extensively addressed in literature \cite{baltruvsaitis2018multimodal,multimodal-repr-learning-survey}. In this work, we focus on the problem of aligning modality-specific representations in a (shared) latent space emerging from the heterogeneity gap between different data sources. Prior work promoting such alignment can be separated into two groups: generation-based methods adjusting Variational Autoencoder (VAE) ~\cite{kingma2013auto} frameworks that considers a prior distribution over the shared latent space, and fusion-based methods that merge modality-specific representations into a shared representation.

\noindent \textbf{Generation-based methods} 
Associative VAE (AVAE)~\citep{yin2017associate} and Joint Multimodal VAE (JMVAE)~\citep{suzuki2016joint} explicitly enforce the alignment of modality-specific representations by minimizing the Kullback–Leibler divergence between their distributions.
However, these models are not easily scalable to large number of modalities due to the combinatorial increase of inference networks required to account for all subsets of modalities. In contrast, GMC scales linearly with the number of modalities as it separately contrasts individual modality-specific representations to the complete ones.

Other multimodal VAE models promote the approximation of modality-specific representations through dedicated training schemes. MVAE~\cite{wu2018multimodal} uses sub-sampling 
to learn a joint-modality representation obtained from a Product-of-Experts (PoE) inference network. This solution is prone to learning overconfident experts, hindering both the alignment of the modality-specific representations and the performance of downstream tasks under incomplete information~\cite{shi2019variational}. Mixture-of-Experts MVAE (MMVAE)~\cite{shi2019variational} instead employs a doubly reparameterized gradient estimator which is computationally expensive compared to the lower-bound objective of traditional multimodal VAEs because of its Monte-Carlo-based training scheme.
GMC, on the other hand, presents an efficient training scheme without suffering from modality-specific biases.

Recently, hierarchical multimodal VAEs have been proposed to facilitate the learning of aligned multimodal representations such as Nexus~\cite{vasco2022leveraging} and Multimodal Sensing (MUSE)~\cite{vasco2021sense}. Nexus considers a two-level hierarchy of modality-specific and multimodal representation spaces employing a dropout-based training scheme. The average aggregator solution employed to merge multimodal information lacks expressiveness which hinders the performance of the model on downstream tasks. To address this issue, MUSE introduces a PoE solution that merges lower-level modality-specific information to encode a high-level multimodal representation, and a dedicated training scheme to counter the overconfident expert issue. In contrast to both solutions, GMC is computationally efficient without requiring hierarchy.

\noindent \textbf{Fusion-based methods}  Other class of methods approach the alignment of modality-specific representations through complex fusion mechanisms \cite{liang2021multibench}. The Multimodal Factorized model (MFM) \cite{tsai2018learning} proposes the factorization of a multimodal representation into distinct multimodal discriminative factors and modality-specific generative factors, which are subsequently fused for downstream tasks. More recently, the Multimodal Transformer model \cite{tsai2019MULT} has shown remarkable classification performance in multimodal time-series datasets, employing a directional pairwise cross-modal attention mechanism to learn a rich representation of heterogeneous data streams without requiring their explicit time-alignment. In contrast to both models, GMC is able to learn multimodal representations of modalities of arbitrary nature without explicitly requiring a supervision signal (e.g. labels).
\section{Experiments}
\label{sec:experiments}

\begin{savenotes}
\begin{table*}[ht!]
\centering
\caption{Performance of different multimodal representation methods in the MHD dataset, in a downstream classification task under complete and partial observations. Accuracy (\%) results averaged over 5 independent runs. Higher is better.}
\vspace{2ex}
\begin{tabular}{@{}lcccccc@{}}
\toprule
 Input & \phantom{1}MVAE\footnotemark & MMVAE & \phantom{1}Nexus & \phantom{1}MUSE & \phantom{1}MFM & \phantom{1}\textbf{GMC (Ours)} \\ \midrule
Complete $(x_{1:4})$ & $100.0 \pm 0.00$  & $99.81 \pm \phantom{0}0.21$ & $99.98 \pm 0.05$  & $99.99 \pm 4\mathrm{e}{-5}$ & $100.0 \pm 0.00$ & $ 100.0 \pm 0.00$ \\ \midrule
Image $(x_{1})$ & $77.94 \pm 3.16$ & $94.63 \pm \phantom{0}2.61$ & $95.89 \pm 0.34$ & $79.37 \pm \phantom{0}2.75$ & $34.66 \pm 6.48$  & $\bf 99.75 \pm 0.03$ \\
Sound $(x_{2})$ & $61.75 \pm 4.59$ & $69.43 \pm 26.43$ & $39.07 \pm 5.82$ & $41.39 \pm \phantom{0}0.18$ & $10.07 \pm 0.20$ & \bf $\bf 93.04 \pm 0.45$ \\
Trajectory $(x_{3})$ & $10.03 \pm 0.06$  & $95.33 \pm \phantom{0}2.56$ & $98.55 \pm 0.34$ & $89.49 \pm \phantom{0}2.44$ & $25.61 \pm 5.41$ & $\bf 99.96 \pm 0.02$ \\
Label $(x_{4})$ & $100.0 \pm 0.00$ & $87.99 \pm \phantom{0}7.49$ & $100.0 \pm 0.00$ & $100.0 \pm \phantom{0}0.00$ & $100.0 \pm 0.00$ & $100.0 \pm 0.00$ \\ \bottomrule
\end{tabular}
\label{table:unsupervised:classification}
\end{table*}
\end{savenotes}

\begin{table*}[t]
\centering
\caption{DCA score of the models in the MHD dataset, evaluating the geometric alignment of complete representations $z_{1:4}$ and modality-specific ones $\{z_1, \dots, z_4\}$ used as $R$ and $E$ inputs in DCA, respectively. The score is averaged over 5 independent runs. Higher is better.}
\vspace{2ex}
\begin{adjustbox}{width=\textwidth,center}
\begin{tabular}{@{}llcccccc@{}}
\toprule
 $R$ & $E$ & \phantom{1}MVAE$^{1}$ & MMVAE & \phantom{1}Nexus & \phantom{1}MUSE & \phantom{1}MFM & \phantom{1}\textbf{GMC (Ours)} \\ \midrule
Complete $(z_{1:4})$ & Image $(z_{1})$ & $0.01 \pm 0.01$ & $0.21 \pm 0.29$ & $0.00 \pm 0.00$ & $0.54 \pm 0.44$ & $0.00 \pm 0.00$ & $\bf 0.96 \pm 0.02$ \\
Complete $(z_{1:4})$ & Sound $(z_{2})$ & $0.00 \pm 0.00$ & $0.00 \pm 0.00$ & $0.00 \pm 0.00$ & $0.00 \pm 0.00$ & $0.00 \pm 0.00$ & $\bf 0.87 \pm 0.16$ \\
Complete $(z_{1:4})$ & Trajectory $(z_{3})$ & $0.00 \pm 0.00$ & $0.01 \pm 0.01$ & $0.08 \pm 0.02$ & $0.00 \pm 0.00$ & $0.00 \pm 0.00$ & $\bf 0.86 \pm 0.05$ \\
Complete $(z_{1:4})$ & Label $(z_{4})$ & $0.99 \pm 0.01$ & $0.74 \pm 0.22$ & $0.43 \pm 0.05$ & $0.93 \pm 0.05$ & $0.85 \pm 0.06$ & $\bf 1.00 \pm 0.00$ \\\bottomrule
\end{tabular}
\end{adjustbox}
\label{table:unsupervised:dca}
\end{table*}

\footnotetext{Results averaged over 3 randomly-seeded runs due to divergence during MVAE training in the remaining seeds.}

We evaluate the quality of the representations learned by GMC on three different scenarios:
\vspace{-0.3cm}
\begin{itemize}
    \item An \emph{unsupervised learning} problem, where we learn multimodal representations on the Multimodal Handwritten Digits (MHD) dataset~\cite{vasco2022leveraging}. We showcase the geometric alignment of representations and demonstrate the superior performance of GMC compared to the baselines on a downstream classification task with missing modalities (Section~\ref{sec:experiments:unsupervised});
     \item A \textit{supervised learning} problem, where we demonstrate the flexibility of GMC by integrating it into state-of-the-art approaches to provide robustness to missing modalities in challenging classification scenarios (Section \ref{sec:experiments:supervised});

    \item A \textit{reinforcement learning} (RL) task, where we show that GMC produces general representations that can be applied to solve downstream control tasks and demonstrate state-of-the-art performance in actuation with missing modality information (Section \ref{sec:experiments:rl}).
\end{itemize}
In each corresponding section, we describe the dataset, baselines, evaluation and training setup used. We report all model architectures and training hyperparameters in Appendix~\ref{sec:appendix:architecture} and~\ref{sec:appendix:hyperparameters}. All results are averaged over $5$ different randomly-seeded runs except for the RL experiments where we consider $10$ runs. Our code is available on GitHub\footnote{\url{https://github.com/miguelsvasco/gmc}}.

\noindent \textbf{Evaluation of geometric alignment} To evaluate the geometric alignment of representations, we use a recently proposed \emph{Delaunay Component Analysis} (DCA)~\citep{dca} method designed for general evaluation of representations. DCA is based on the idea of comparing geometric and topological properties of an evaluation set of representations $E$ with the reference set $R$, acting as an approximation of the true underlying manifold. The set $E$ is considered to be well aligned with $R$ if its global and local structure resembles well the one captured by $R$, i.e., the manifolds described by the two sets have similar number, structure and size of connected components.

DCA approximates the manifolds described by $R$ and $E$ with a Delaunay neighbourhood graph and derives several scores reflecting their alignment. We consider three of them: 
\emph{network quality} $q \in [0, 1]$ which measures the overall geometric alignment of $R$ and $E$ in the connected components, as well as \emph{precision} $\mathcal{P} \in [0, 1]$ and \emph{recall} $\mathcal{R} \in [0, 1]$ which measure the proportion of points from $E$ and $R$, respectively, that are contained in geometrically well-aligned components. To account for all three normalized scores, we report the harmonic mean defined as $3/ (1/\mathcal{P} + 1/\mathcal{R} + 1/q)$ when all $\mathcal{P}, \mathcal{R}, q > 0$ and $0$ otherwise. In all experiments, we compute DCA using complete representations $z_{1:M}$ as the reference set $R$ and modality-specific $z_m$ as the evaluation set $E$, both obtained from testing observations. A detailed description of the method and definition of the scores is found in Appendix \ref{sec:appendix:dca}.

\subsection{Experiment 1: Unsupervised Learning}
\label{sec:experiments:unsupervised}

\textbf{Datasets} The MHD dataset is comprised of images ($x_1$), sounds ($x_2$), motion trajectories ($x_3$) and label information ($x_4$) related to handwriting digits. The authors collected $60,000$ $28 \times 28$ greyscale images per class as well as normalized $200$-dimensional representations of trajectories and $128 \times 32$-dimensional representations of audio. The dataset is split into $50,000$ training and $10,000$ testing samples.

\begin{table}[t]
 \caption{Number of parameters (in millions) of the representation models employed in the Multimodal Handwritten Digits dataset.}
     \centering
     \vspace{2ex}
     \begin{adjustbox}{width=\columnwidth,center}
    \begin{tabular}{@{}cccccc@{}}
\toprule
MVAE & MMVAE & Nexus & MUSE & MFM & \textbf{GMC (Ours)} \\ \midrule
9.3 & 9.0 & 12.9 & 9.9 & 9.4 & \textbf{2.9} \\ \bottomrule
\end{tabular}
\end{adjustbox}
\label{table:unsupervised:complexity}
\vspace{-1ex}
\end{table}

\textbf{Models} We consider several generation-based and fusion-based state-of-the-art multimodal representation methods:  MVAE, MMVAE, Nexus, MUSE and MFM (see Section \ref{section:related_work} for a detailed description). For a fair comparison, when possible, we employ the same encoder architectures and latent space dimensionality across all baseline models, described in Appendix \ref{sec:appendix:architecture}. For GMC, we employ the same modality-specific base encoders $f_m(\cdot)$ as the baselines with an additional base encoder $f_{1:4}(\cdot)$ taking complete observations as input. 
The shared projection head $g(\cdot)$ comprises of $3$ fully-connected layers. 
We set the temperature $\tau=0.1$ and consider $64$-dimensional intermediate and shared representation spaces, i.e., $h \in \mathbb{R}^{64}, z \in \mathbb{R}^{64}$. We train all the models for 100 epochs using a learning rate of $10^{-3}$, employing the training schemes and hyperparameters suggested by the authors (when available). 

\begin{table*}[t]
     \caption{Performance of different multimodal representation methods in the CMU-MOSEI dataset, in a classification task under complete and partial observations. Results averaged over 5 independent runs. Arrows indicate the direction of improvement.}
     \centering
     \vspace{2ex}
    \begin{subtable}[h]{0.45\textwidth}
        \centering
        \begin{tabular}{@{}lcc@{}}
        \toprule
        Metric & Baseline & GMC (Ours)  \\ \midrule
        MAE ($\downarrow$) & $0.643 \pm 0.019$   & $\bf 0.634 \pm 0.008$  \\
        Cor ($\uparrow$) & $\bf 0.664 \pm 0.004$  & $0.653 \pm 0.004$  \\
        F1  ($\uparrow$)& $\bf 0.809 \pm 0.003$  & $0.798 \pm 0.008$  \\
        Acc ($\%, \uparrow$)& $\bf 80.75 \pm 00.28$  & $79.73 \pm 00.69$  \\ \bottomrule
        \end{tabular}
       \caption{Complete Observations $(x_{1:3})$}
       \label{table:supervised:mosei:complete}
    \end{subtable}
    \hfill
    \begin{subtable}[h]{0.45\textwidth}
        \centering
        \begin{tabular}{@{}lcc@{}}
        \toprule
        Metric & Baseline & \textbf{GMC (Ours)} \\ \midrule
        MAE ($\downarrow$) & $0.805 \pm 0.028$   & $\bf 0.712 \pm 0.015$  \\
        Cor ($\uparrow$) & $0.427 \pm 0.061$  & $\bf 0.590 \pm 0.013$  \\
        F1  ($\uparrow$)& $0.713 \pm 0.086$  & $\bf 0.779 \pm 0.005$  \\
        Acc ($\%, \uparrow$)& $66.53 \pm 09.86$  & $\bf 77.85 \pm 00.36$  \\ \bottomrule
        \end{tabular}
        \caption{Text Observations $(x_1)$}
        \label{table:supervised:mosei:text}
     \end{subtable}
     
      \vspace{1ex}
     
     \begin{subtable}[h]{0.45\textwidth}
        \centering
        \begin{tabular}{@{}lcc@{}}
        \toprule
        Metric & Baseline & \textbf{GMC (Ours)} \\ \midrule
        MAE ($\downarrow$) & $0.873 \pm 0.065$   & $\bf 0.837 \pm 0.008$  \\
        Cor ($\uparrow$) & $0.090 \pm 0.062$  & $\bf 0.256 \pm 0.007$  \\
        F1  ($\uparrow$)& $0.622 \pm 0.122$  & $\bf 0.676 \pm 0.015$  \\
        Acc ($\%, \uparrow$)& $53.17 \pm 09.47$  & $\bf 65.59 \pm 00.62$  \\ \bottomrule
        \end{tabular}
        \caption{Audio Observations $(x_2)$}
        \label{table:supervised:mosei:audio}
     \end{subtable}
    \hfill
    \begin{subtable}[h]{0.45\textwidth}
        \centering
        \begin{tabular}{@{}lcc@{}}
        \toprule
        Metric & Baseline & \textbf{GMC (Ours)} \\ \midrule
        MAE ($\downarrow$) & $1.025 \pm 0.164$   & $\bf 0.845 \pm 0.010$  \\
        Cor ($\uparrow$) & $0.110 \pm 0.060$  & $\bf 0.278 \pm 0.011$  \\
        F1  ($\uparrow$)& $0.574 \pm 0.095$  & $\bf 0.655 \pm 0.003$  \\
        Acc ($\%, \uparrow$)& $44.33 \pm 09.40$  & $\bf 65.02 \pm 00.28$  \\ \bottomrule
        \end{tabular}
        \caption{Video Observations $(x_3)$}
        \label{table:supervised:mosei:video}
     \end{subtable}
     \label{table:supervised:mosei}
\end{table*}

\textbf{Evaluation} We follow the established evaluation in the literature using classification as a downstream task \cite{shi2019variational} and train a $10$-class classifier neural network on complete representations $z_{1:M} = g\left(f_{1:M}(x_{1:M})\right)$ from the training split (see Appendix \ref{sec:appendix:architecture} for the exact architecture). The classifier is trained for 50 epochs using a learning rate of $1\mathrm{e}{-3}$. We report the testing accuracy obtained when the classifier is provided with both complete $z_{1:4}$ and modality-specific representations $z_{m}$ as inputs.

\textbf{Classification results} The classification results are shown in Table~\ref{table:unsupervised:classification}. While all the models attain perfect accuracy on $x_{1:4}$ and $x_4$, we observe that GMC is the only model that successfully performs the task when given only $x_1, x_2$ or $x_3$ as input,
significantly outperforming the baselines.

\comment{
\begin{table*}[t]
     \caption{Performance of different multimodal representation methods in the CMU-MOSEI dataset, in a classification task under complete and partial observations. Results averaged over 5 independent runs. Arrows indicate the direction of improvement.}
     \centering
     \vspace{2ex}
    \begin{subtable}[h]{0.45\textwidth}
        \centering
        \begin{tabular}{@{}lcc@{}}
        \toprule
        Metric & Baseline & GMC (Ours)  \\ \midrule
        MAE ($\downarrow$) & $0.643 \pm 0.019$   & $\bf 0.634 \pm 0.008$  \\
        Cor ($\uparrow$) & $\bf 0.664 \pm 0.004$  & $0.653 \pm 0.004$  \\
        F1  ($\uparrow$)& $\bf 0.809 \pm 0.003$  & $0.798 \pm 0.008$  \\
        Acc ($\%, \uparrow$)& $\bf 80.75 \pm 00.28$  & $79.73 \pm 00.69$  \\ \bottomrule
        \end{tabular}
       \caption{Complete Observations $(x_{1:3})$}
       \label{table:supervised:mosei:complete}
    \end{subtable}
    \hfill
    \begin{subtable}[h]{0.45\textwidth}
        \centering
        \begin{tabular}{@{}lcc@{}}
        \toprule
        Metric & Baseline & \textbf{GMC (Ours)} \\ \midrule
        MAE ($\downarrow$) & $0.805 \pm 0.028$   & $\bf 0.712 \pm 0.015$  \\
        Cor ($\uparrow$) & $0.427 \pm 0.061$  & $\bf 0.590 \pm 0.013$  \\
        F1  ($\uparrow$)& $0.713 \pm 0.086$  & $\bf 0.779 \pm 0.005$  \\
        Acc ($\%, \uparrow$)& $66.53 \pm 09.86$  & $\bf 77.85 \pm 00.36$  \\ \bottomrule
        \end{tabular}
        \caption{Text Observations $(x_1)$}
        \label{table:supervised:mosei:text}
     \end{subtable}
     
      \vspace{1ex}
     
     \begin{subtable}[h]{0.45\textwidth}
        \centering
        \begin{tabular}{@{}lcc@{}}
        \toprule
        Metric & Baseline & \textbf{GMC (Ours)} \\ \midrule
        MAE ($\downarrow$) & $0.873 \pm 0.065$   & $\bf 0.837 \pm 0.008$  \\
        Cor ($\uparrow$) & $0.090 \pm 0.062$  & $\bf 0.256 \pm 0.007$  \\
        F1  ($\uparrow$)& $0.622 \pm 0.122$  & $\bf 0.676 \pm 0.015$  \\
        Acc ($\%, \uparrow$)& $53.17 \pm 09.47$  & $\bf 65.59 \pm 00.62$  \\ \bottomrule
        \end{tabular}
        \caption{Audio Observations $(x_2)$}
        \label{table:supervised:mosei:audio}
     \end{subtable}
    \hfill
    \begin{subtable}[h]{0.45\textwidth}
        \centering
        \begin{tabular}{@{}lcc@{}}
        \toprule
        Metric & Baseline & \textbf{GMC (Ours)} \\ \midrule
        MAE ($\downarrow$) & $1.025 \pm 0.164$   & $\bf 0.845 \pm 0.010$  \\
        Cor ($\uparrow$) & $0.110 \pm 0.060$  & $\bf 0.278 \pm 0.011$  \\
        F1  ($\uparrow$)& $0.574 \pm 0.095$  & $\bf 0.655 \pm 0.003$  \\
        Acc ($\%, \uparrow$)& $44.33 \pm 09.40$  & $\bf 65.02 \pm 00.28$  \\ \bottomrule
        \end{tabular}
        \caption{Video Observations $(x_3)$}
        \label{table:supervised:mosei:video}
     \end{subtable}
     \label{table:supervised:mosei}
\end{table*}
}

\textbf{Geometric alignment} To validate that the superior performance of GMC originates from a better geometric alignment of representations, we evaluate the testing representations obtained from all the models using DCA. For each modality $m$, we compared the alignment of the evaluation set $E = \{ z_m \}$ and the reference set $R = \{z_{1:4}\}$. The obtained DCA scores are shown in Table~\ref{table:unsupervised:dca} where we see that GMC outperforms all the considered baselines. For some cases, we observe the obtained representations are completely misaligned yielding $\mathcal{P} = \mathcal{R} = q = 0$. While some of the baselines are to some extend able to align $z_1$ and/or $z_4$ to $z_{1:4}$, GMC is the only method that is able to align even the sound and trajectory representations, $z_2$ and $z_3$, resulting in a superior classification performance.

We additionally validate the geometric alignment by visualizing 2-dimensional UMAP projections~\cite{UMAP} of the representations $z$. In Figure~\ref{fig:misalignment} we show projections of $z_{1:4}$ and image representations $z_1$ obtained using the considered models. We clearly see that GMC not only correctly aligns $z_{1:4}$ and $z_1$ but also separates the representations in $10$ clusters. Moreover, we can see that among the baselines only MMVAE and MUSE somewhat align the representations which is on par with the quantitative results reported in Table~\ref{table:unsupervised:dca}. For MVAE, Nexus and MFM, Figure~\ref{fig:misalignment} visually supports the obtained DCA score $0$. Note that points marked as outliers by DCA are omited from the visualization. We provide similar visualizations of other modalities in Appendix \ref{sec:appendix:vis}.

\textbf{Model Complexity} In Table \ref{table:unsupervised:complexity} we present the number of parameters required by the multimodal representation models employed in this task. The results show that GMC requires significantly fewer parameters than the smallest baseline model -- 68\% fewer parameters than MMVAE.

\comment{
\begin{table}[h]
 \caption{Number of parameters (in millions) of the representation models employed in the Multimodal Handwritten Digits dataset.}
     \centering
     \vspace{2ex}
     \begin{adjustbox}{width=\columnwidth,center}
    \begin{tabular}{@{}cccccc@{}}
\toprule
MVAE & MMVAE & Nexus & MUSE & MFM & \textbf{GMC (Ours)} \\ \midrule
9.3 & 9.0 & 12.9 & 9.9 & 9.4 & \textbf{2.9} \\ \bottomrule
\end{tabular}
\end{adjustbox}
\label{table:unsupervised:complexity}
\end{table}
}

\begin{table}[t]
\centering
\caption{DCA score of the models in the CMU-MOSEI dataset evaluating the geometric alignment of complete representations $z_{1:4}$ and modality-specific ones $\{z_1, z_2, z_3\}$ used as $R$ and $E$ inputs in DCA, respectively. The score is averaged over 5 independent runs. Higher is better.}
\vspace{2ex}
  \begin{adjustbox}{width=\columnwidth,center}
\begin{tabular}{@{}lccc@{}}
\toprule
\multicolumn{1}{c}{R} & \multicolumn{1}{c}{E} & \multicolumn{1}{c}{Baseline} & \multicolumn{1}{c}{\textbf{GMC (Ours)}} \\ \midrule

Complete $(z_{1:3})$ & Text $(z_{1})$ & $0.50 \pm 0.05$ & $\bf 0.95 \pm 0.01$  \\
Complete $(z_{1:3})$ & Audio $(z_{2})$ & $0.41 \pm 0.14$ & $\bf 0.86 \pm 0.04$ \\
Complete $(z_{1:3})$ & Vision $(z_{3})$ & $0.50 \pm 0.14$ & $\bf 0.92 \pm 0.02$  \\\bottomrule
\end{tabular}
\end{adjustbox}
\label{table:supervised:dca}
\end{table}
\vspace{-0.3cm}

\subsection{Experiment 2: Supervised Learning}
\label{sec:experiments:supervised}
In this section, we evaluate the flexibility of GMC by adjusting both the architecture of the model and training procedure to receive an additional supervision signal during training to guide the learning of complete representations. We demonstrate how GMC can be integrated into existing approaches to provide additional robustness to missing modalities with minimal computational cost.

\textbf{Datasets} We employ the CMU-MOSI~\citep{mosi} and CMU-MOSEI~\citep{mosei}, two popular datasets for sentiment analysis and emotion recognition with challenging temporal dynamics. Both datasets consist of textual ($x_1$), sound ($x_2$) and visual ($x_3$) modalities extracted from videos. CMU-MOSI consists of 2199 short monologue videos clips of subjects expressing opinions about various topics. CMU-MOSEI is an extension of CMU-MOSI dataset containing 23453 YouTube video clips of subjects expressing movie reviews. In both datasets, each video clip is annotated with labels in $[-3, 3]$, where ${-3}$ and $3$ indicate strong negative and strongly positive sentiment scores, respectively. We employ the temporally-aligned version of these datasets: CMU-MOSEI consists of $18134$ and $4643$ training and testing samples, respectively, and  CMU-MOSI consists of $1513$ and $686$ training and testing samples, respectively.

\textbf{Models} We consider the Multimodal Transformer \cite{tsai2019MULT} which is the state-of-the-art model for classification on the CMU-MOSI and CMU-MOSEI datasets (we refer to \citet{tsai2019MULT} for a detailed description of the architecture). For GMC, we employ the same architecture for the joint-modality encoder $f_{1:3}(\cdot)$ as the Multimodal Transformer but remove the last classification layers. For the modality-specific base encoders $\{f_1(\cdot), f_2(\cdot), f_3(\cdot)\}$, we employ a simple GRU layer with 30 hidden units and a fully-connected layer. The shared projection head $g(\cdot)$ is comprised of a single fully connected layer. We set $\tau = 0.3$ and consider $60$-dimensional intermediate and shared representations $h, z \in \mathbb{R}^{60}$.

In addition, we employ a simple classifier consisting of 2 linear layers over the complete representations $z_{1:M}$ to provide the supervision signal to the model during training. We follow the training scheme proposed by \citet{tsai2019MULT} and train all models for 40 epochs with a decaying learning rate of $10^{-3}$.

\textbf{Evaluation} We evaluate the performance of representation learning models in sentiment analysis classification with missing modality information. We consider the same metrics as in \citet{tsai2018learning,tsai2019MULT} and report binary accuracy (Acc), mean absolute error (MAE), correlation (Cor) and F1 score (F1) of the predictions obtain on the test dataset. In Appendix \ref{sec:appendix:mosi} we present similar results on the CMU-MOSI dataset.

\begin{table*}[t]
\centering
\caption{Performance after zero-shot policy transfer in the multimodal Pendulum task. At test time, the agent is provided with either image ($x_{1}$), sound ($x_{2}$), or complete ($x_{1:2}$) observations. Total reward averaged over 100 episodes and 10 randomly seeded runs. Higher is better.}
\vspace{2ex}
\begin{tabular}{@{}lccc@{}}
\toprule
       Observation & MVAE + DDPG & MUSE + DDPG & \textbf{GMC + DDPG (Ours)}     \\ \midrule
Complete $(x_{1:2})$ & $-1.114 \pm 0.110$ & $-1.005 \pm 0.117$ & $\bf -0.935 \pm 0.057$ \\
Image $(x_{1})$ &  $ -1.116 \pm 0.121$  & $-4.752 \pm 0.994$ & $\bf -0.940 \pm 0.056$ \\
Sound $(x_2)$ & $-6.642 \pm 0.106$ & $-3.459 \pm 0.519$ & $\bf-0.956 \pm 0.075$ \\ \bottomrule
\end{tabular}
\label{table:rl:pendulum}
\end{table*}

\begin{table*}[t]
\centering
\caption{DCA score of the models in the multimodal Pendulum task evaluating the geometric alignment of complete representations $z_{1:2}$ and modality-specific ones $\{z_1, z_2\}$ used as $R$ and $E$ inputs in DCA, respectively. Results averaged over 10 independent runs. Higher is better.}
\vspace{2ex}
\begin{tabular}{@{}llccc@{}}
\toprule
       $R$ & $E$ & MVAE + DDPG & MUSE + DDPG & \textbf{GMC + DDPG (Ours)}     \\ \midrule
Complete $(z_{1:2})$ & Image $(z_1)$ & $0.79 \pm 0.01$ & $0.20 \pm 0.09$ & $\bf 0.87 \pm 0.01$ \\
Complete $(z_{1:2})$ & Sound $(z_2)$ & $0.00 \pm 0.00$ & $0.01 \pm 0.01$ & $\bf 0.88 \pm 0.02$ \\
\bottomrule
\end{tabular}
\label{table:rl:pendulum_dca}
\end{table*}

\textbf{Results} The results obtained on CMU-MOSEI are reported in Table~\ref{table:supervised:mosei}. When using the complete observations $x_{1:3}$ as inputs, GMC achieves competitive performance with the baseline model indicating that the additional contrastive loss does not deteriorate the model's capabilities (Table~\ref{table:supervised:mosei:complete}). 
However, GMC significantly improves the robustness of the  model to the missing modalities as seen in Tables~\ref{table:supervised:mosei:text},~\ref{table:supervised:mosei:audio} and~\ref{table:supervised:mosei:video} where we use only individual modalities as inputs. 
While GMC consistently outperforms the baseline in all metrics, we observe the largest improvement on the F1 score and binary accuracy (Acc) where the baseline often performs worse than random. As before, we additionally evaluate the geometric alignment of the modality-specific representations $z_m$ (comprising the set $E$) and complete representations $z_{1:3}$ (comprising the set $R$). The resulting DCA score, reported in Table \ref{table:supervised:dca}, supports the results shown in Table \ref{table:supervised:mosei} and verifies that GMC significantly improves the geometric alignment compared to the baseline. Furthermore, GMC incurs in a small computational cost (with 1.4 million parameters), requiring only 300K extra parameters in comparison with the baseline (with 1.1 million parameters).

\subsection{Experiment 3: Reinforcement Learning}
\label{sec:experiments:rl}

\begin{table}[t]
 \caption{Number of parameters (in millions) of the representation models employed in the multimodal Pendulum scenario.}
     \centering
     \vspace{2ex}
    \begin{tabular}{@{}ccc@{}}
\toprule
MVAE & MUSE & \textbf{GMC (Ours)} \\ \midrule
3.8  & 4.3  & \textbf{1.9} \\ \bottomrule
\end{tabular}
\label{table:rl:complexity}
\end{table}

In this section, we demonstrate how GMC can be employed as a representation model in the design of RL agents yielding state-of-the-art performance using missing modality information during task execution.

\textbf{Scenario} We consider the recently proposed multimodal inverted Pendulum task~\cite{silva2019playing} which is an extension of the classical control scenario to a multimodal setting. In this task, the goal is to swing the pendulum up so it remains balanced upright. The observations of the environment include both an image ($x_1$) and a sound ($x_2$) component. The sound component is generated by the tip of the pendulum emitting a constant frequency $f_0$. This frequency is received by a set of $S$ sound receivers $\left\lbrace \rho_1, \dots, \rho_S \right\rbrace$. At each timestep, the frequency $f'_i$ heard by each sound receiver $\rho_i$ is modified by the Doppler effect, modifying the frequency heard by an observer as a function of the velocity of the sound emitter. The amplitude is modified as function of the relative position of the emitter in relation to the observer following an inverse square law. To train the representation models, we employ a random policy to collect a dataset composed of 20,000 training samples and 2,000 test samples following the procedure of \citet{silva2019playing}. 

\textbf{Models} We consider the MVAE~\cite{wu2018multimodal} and the MUSE~\cite{vasco2021sense} models which are two commonly used approaches for the perception of multimodal RL agents. For GMC, we employ the same modality-specific encoders $f_1(\cdot), f_2(\cdot)$ as the baselines in addition to a joint-modality encoder $f_{1:2}(\cdot)$. The shared projection head $g(\cdot)$ is comprised of $2$ fully-connected layers. 
We use $\tau = 0.3$ and set the dimensions of intermediate and latent representations spaces to $d = 64$ and $s = 10$. We follow the two-stage agent pipeline proposed in \citet{higgins2017darla} and initially train all representation models on the dataset of collected observations for 500 epochs using a learning rate of $10^{-3}$. We subsequently train a  Deep Deterministic Policy Gradient (DDPG) controller \cite{lillicrap2015continuous} that takes as input the representations $z_{1:2}$ encoded from complete observations $x_{1:2}$ following the network architecture and training hyperparameters used by \citet{silva2019playing}.

\textbf{Evaluation} We evaluate the performance of RL agents acting under incomplete perceptions that employ the representation models to encode raw observations of the environment. During execution, the environment may provide any of the modalities \{$x_{1:2}, x_1, x_2 $\}. As such, we compare the performance of the RL agents when directly using the policy learned from complete observations in scenarios with possible missing modalities without any additional training (zero-shot transfer).

\textbf{Results}
Table \ref{table:rl:pendulum} summarizes the total reward collected per episode for the Pendulum scenario averaged over 100 episodes and 10 randomly seeded runs\footnote{Results averaged over 9 randomly-seeded runs for the MUSE + DDPG method due to divergence during training in the remaining seed.}. 

The results show that only GMC is able to provide the agent with a representation model robust to partial observations allowing the agent to act under incomplete perceptual conditions with no performance loss. 
This is on par with the DCA scores reported in Table~\ref{table:rl:pendulum_dca} indicating that GMC geometrically better aligns the representations compared to the baselines. Once again, as shown in Table \ref{table:rl:complexity}, GMC can achieve such performance with $50\%$ fewer parameters than the smallest baseline, evidence of its efficiency.

\subsection{Ablation studies}
We perform an ablation study on the hyperparameters of GMC using the setup from Section~\ref{sec:experiments:unsupervised} on MHD dataset. In particular, we investigate: \textit{i)} the robustness of the GMC framework when varying the temperature parameter $\tau$;  \textit{ii)} the performance of GMC when varying dimensionalities $d$ and $s$ of the intermediate and latent representation spaces, respectively; and \textit{iii)} the performance of GMC trained with a modified loss function that uses only complete observations as negative pairs. We report both classification results and DCA scores in Appendix \ref{section:appendix:ablation} and observe that GMC is robust to different experimental conditions both in terms of performance and geometric alignment of representations.
\section{Conclusion}
We addressed the problem of learning multimodal representations that are both semantically rich and robust to missing modality information. We contributed with a novel Geometric Multimodal Contrastive (GMC) learning framework that is inspired by the visual contrastive learning methods and geometrically aligns complete and modality-specific representations in a shared latent space. We have shown that GMC is able to achieve state-of-the-art performance with missing modality information across a wide range of different learning problems while being computationally efficient (often requiring 90\% fewer parameters than similar models) and straightforward to integrate with existing state-of-the-art approaches. We believe that GMC broadens the range of possible applications of contrastive learning methods to multimodal scenarios and opens many future work directions, such as investigating the effect of modality-specific augmentations or usage of inherent intermediate representations for modality-specific downstream tasks.

\section*{Acknowledgements} 
This work has been supported by the Knut and Alice Wallenberg Foundation, Swedish Research Council and European Research Council. This work was also partially supported by Portuguese national funds through the Portuguese Fundação para a Ciência e a Tecnologia under project UIDB/50021/2020 (INESC-ID multi annual funding) and project PTDC/CCI-COM/5060/2021. In addition, this research was partially supported by TAILOR, a project funded by EU Horizon 2020 research and innovation programme under GA No. 952215. This work was also supported by funds from Europe Research Council under project BIRD 884887. Miguel Vasco acknowledges the Fundação para a Ciência e a Tecnologia PhD grant SFRH/BD/139362/2018.

\bibliography{references}
\bibliographystyle{icml2022}

\clearpage
\appendix
\onecolumn

\section{Delaunay Component Analysis}
\label{sec:appendix:dca}

Delaunay Component Analysis (DCA) is a recently proposed method for general evaluation of data representations~\cite{dca}. The basic idea of DCA is to compare geometric and topological properties of two sets of representations -- a reference set $R$ representing the true underlying data manifold and an evaluation set $E$. If the sets $R$ and $E$ represent data from the same underlying manifold, then the geometric and topological properties extracted from manifolds described by $R$ and $E$ should be similar. DCA approximates these manifolds using a type of a neighbourhood graph called Delaunay graph $\mathcal{G}$ build on the union $R \cup E$. The alignment of $R$ and $E$ is then determined by analysing the connected components of $\mathcal{G}$ from which several global and local scores are derived. 

DCA first evaluates each connected component $\mathcal{G}_i$ of $\mathcal{G}$ by analyzing the number of points from $R$ and $E$ contained in $\mathcal{G}_i$ as well as number of edges among these points. In particular, each component $\mathcal{G}_i$ is evaluated by two scores: \textit{consistency} and \textit{quality}. Intuitively, $\mathcal{G}_i$ has a high consistency if it is equality represented by points from $R$ and $E$, and high quality if $R$ and $E$ points are geometrically well aligned. The latter holds true if the number of homogeneous edges among points in each of the sets is small compared to the number of heterogeneous edges connecting representations from $R$ and $E$.

To formally define the scores, we follow \citet{dca}: for a graph $\mathcal{G} = (\mathcal{V}, \mathcal{E})$ we denote by $|\mathcal{G}|_\mathcal{V}$ the size of its vertex set and by $|\mathcal{G}|_\mathcal{E}$ the size of its edge set. Moreover, $\mathcal{G}^Q = (\mathcal{V}|_Q, \mathcal{E}|_{Q \times Q}) \subset \mathcal{G}$ denotes its restriction to a set $Q \subset \mathcal{V}$. 

\begin{definition} \label{def:local_scores}
Consistency $c$ and quality $q$ of a connected component $\mathcal{G}_i \subset \mathcal{G}$ are defined as the ratios
\begin{align*}
    c(\mathcal{G}_i) &= 1 - \frac{|\, |\mathcal{G}_i^R|_\mathcal{V} - |\mathcal{G}_i^{E}|_\mathcal{V} \,|}{|\mathcal{G}_i|_\mathcal{V}}, \\
    q(\mathcal{G}_i) &= 
    \begin{cases}
    1 -  \frac{(| \mathcal{G}_i^{R}|_\mathcal{E} + |\mathcal{G}_i^{E}|_\mathcal{E})}{|\mathcal{G}_i|_\mathcal{E}} & \text{if } |\mathcal{G}_i|_\mathcal{E} \geq 1\\
    0              & \text{otherwise},
\end{cases}
\end{align*}
respectively. Moreover, the scores computed on the entire Delaunay graph $\mathcal{G}$ are called network consistency $c(\mathcal{G})$ and network quality $q(\mathcal{G})$.
\end{definition}

Besides the two global scores, \textit{network consistency} and \textit{network quality} defined above,  two more global similarity scores are derived from the local ones by extracting the so-called \textit{fundamental} components of high consistency and high quality. In this work, we define a component $\mathcal{G}_i$ to be fundamental if $c(\mathcal{G}_i) > 0$ and $q(\mathcal{G}_i) > 0$ and denote by $\mathcal{F} $ the union of all fundamental components of the Delaunay graph $\mathcal{G}$. By examining the proportion of points from $E$ and $R$ that are contained in $\mathcal{F}$, DCA derives two global scores \textit{precision} and \textit{recall} defined below.

\begin{definition}
Precision $\mathcal{P}$ and recall $\mathcal{R}$ associated to a Delaunay graph $\mathcal{G}$ built on $R \cup E$ are defined as 
\begin{align*}
    \mathcal{P} = \frac{|\mathcal{F}^E|_\mathcal{V}}{|\mathcal{G}^E|_\mathcal{V}} \quad \text{and} \quad \mathcal{R} = \frac{|\mathcal{F}^R|_\mathcal{V}}{|\mathcal{G}^R|_\mathcal{V}},
\end{align*}
respectively, 
where $\mathcal{F}^R, \mathcal{F}^E$ are the restrictions of $\mathcal{F}$ to the sets $R$ and $E$.
\end{definition}

We refer the reader to~\citet{dca,geomca} for further details.

\section{Ablation Study on GMC}
\label{section:appendix:ablation}

We perform a ablation study on the hyperparameters of GMC using the setup from Section~\ref{sec:experiments:unsupervised} on the MHD dataset. In particular, we investigate:
\begin{enumerate}
    \item the robustness of the GMC framework when varying the temperature parameter $\tau$;
    \item the performance of GMC with different dimensionalities of the intermediate representations  $h \in \mathbb{R}^d$;
     \item the performance of GMC with different dimensionalities of the shared latent representations $z \in \mathbb{R}^s$;
     \item the performance of GMC with a modified loss $\mathcal{L}_{\text{GMC}}^{*}$ that only uses complete observations as negative pairs.
\end{enumerate}

In all experiments we report both classification results and DCA scores.

\newpage

\begin{table}[H]
\centering
\caption{Performance of GMC with different temperature values $\tau$ (Equation \eqref{eq:similarity}) in the MHD dataset, in a downstream classification task under complete and partial observations. Accuracy results averaged over 5 independent runs. Higher is better.}
\vspace{2ex}
\begin{tabular}{@{}lccccc@{}}
\toprule
Observations & $\tau = 0.05$ & $\tau = 0.1$ (Default) & $\tau = 0.2$ & $\tau = 0.3$ & $\tau = 0.5$   \\ \midrule
Complete Observations & $\phantom{1}99.99 \pm 0.01$  & $100.00 \pm 0.00$  & $\phantom{1}99.99 \pm 0.01$  & $\phantom{1}99.97 \pm 3\mathrm{e}{-5}$  & $\phantom{1}99.96 \pm 0.01$   \\  \midrule
Image Observations & $\phantom{1}99.78 \pm 0.02$  & $\phantom{1}99.75 \pm 0.03$  & $\phantom{1}99.84 \pm 0.03$  & $\phantom{1}99.80 \pm 0.04$  & $\phantom{1}99.89 \pm 0.03$   \\
Sound Observations & $\phantom{1}93.55 \pm 0.22$  & $\phantom{1}93.04 \pm 0.45$  & $\phantom{1}91.98 \pm 0.29$  & $\phantom{1}91.87 \pm 0.58$  & $\phantom{1}95.01 \pm 0.38$  \\
Trajectory Observations & $\phantom{1}99.94 \pm 0.01$ & $\phantom{1}99.96 \pm 0.02$  & $\phantom{1}99.97 \pm 0.02$  & $\phantom{1}99.96 \pm 0.01$   & $\phantom{1}99.80 \pm 0.20$   \\
Label Observations & $100.00 \pm 0.00$ & $100.00 \pm 0.00$  & $100.00 \pm 0.00$  & $100.00 \pm 0.00$  & $100.00 \pm 0.00$  \\ \bottomrule
\end{tabular}
\label{table:ablation:temperature:classification}
\vspace{-4ex}
\end{table}

\begin{table}[H]
\centering
\caption{DCA score obtained on GMC representations when trained with different temperature values $\tau$ (Equation \eqref{eq:similarity}) in the MHD dataset, evaluating the geometric alignment of complete representations $z_{1:4}$ and modality-specific ones $\{z_1, \dots, z_4\}$ used as $R$ and $E$ inputs in DCA, respectively. The score is averaged over 5 independent runs. Higher is better.}
\vspace{2ex}
\begin{tabular}{@{}llccccc@{}}
\toprule
 $R$ & $E$ & $\tau = 0.05$ & $\tau = 0.1$  (Default) & $\tau = 0.2$ & $\tau = 0.3$ & $\tau = 0.5$ \\ \midrule
Complete $(z_{1:4})$ & Image $(z_{1})$ & $0.96 \pm 0.02$ & $0.96 \pm 0.02$ & $0.93 \pm 0.01$ & $0.92 \pm 0.00$ & $0.89 \pm 0.02$  \\
Complete $(z_{1:4})$ & Sound $(z_{2})$ & $0.95 \pm 0.02$ & $0.87 \pm 0.16$ & $0.96 \pm 0.02$ & $0.99 \pm 0.00$ & $0.87 \pm 0.04$ \\
Complete $(z_{1:4})$ & Trajectory $(z_{3})$ & $0.96 \pm 0.02$ & $0.86 \pm 0.05$ & $0.90 \pm 0.03$ & $0.92 \pm 0.00$ & $0.64 \pm 0.11$ \\
Complete $(z_{1:4})$ & Label $(z_{4})$ & $1.00 \pm 0.00$ & $1.00 \pm 0.00$ & $1.00 \pm 0.00$ & $1.00 \pm 0.00$ & $0.94 \pm 0.02$ \\\bottomrule
\end{tabular}
\label{table:ablation:temperature:dca}
\vspace{-4ex}
\end{table}

\begin{table}[H]
\centering
\caption{Performance of GMC with different values of intermediate representation dimensionality $h \in \mathbb{R}^d$ in the MHD dataset, in a downstream classification task under complete and partial observations. Accuracy results averaged over 5 independent runs. Higher is better.}
\vspace{2ex}
\begin{tabular}{@{}lccc@{}}
\toprule
Observations & $d = 32$ & $d = 64$ (Default) & $d = 128$   \\ \midrule
Complete Observations & $\phantom{1}99.99 \pm 0.01$  & $100.00 \pm 0.00$  & $\phantom{1}99.99 \pm 0.01$  \\  \midrule
Image Observations & $\phantom{1}99.75 \pm 0.04$  & $\phantom{1}99.75 \pm 0.03$  & $\phantom{1}99.72 \pm 0.07$     \\
Sound Observations & $\phantom{1}93.31 \pm 0.41$  & $\phantom{1}93.04 \pm 0.45$  & $\phantom{1}93.34 \pm 0.51$   \\
Trajectory Observations & $\phantom{1}99.96 \pm 0.01$  & $\phantom{1}99.96 \pm 0.02$  & $\phantom{1}99.96 \pm 0.01$    \\
Label Observations & $100.00 \pm 0.00$  & $100.00 \pm 0.00$  & $100.00 \pm 0.00$  \\ \bottomrule
\end{tabular}
\label{table:ablation:intermediate:classification}
\vspace{-4ex}
\end{table}

\begin{table}[H]
\centering
\caption{DCA score obtained on GMC representations when varying the dimension of intermediate representations $h \in \mathbb{R}^d$ in the MHD dataset, evaluating the geometric alignment of complete representations $z_{1:4}$ and modality-specific ones $\{z_1, \dots, z_4\}$ used as $R$ and $E$ inputs in DCA, respectively. The score is averaged over 5 independent runs. Higher is better.}
\vspace{2ex}
\begin{tabular}{@{}llccc@{}}
\toprule
 $R$ & $E$ & $d = 32$ & $d = 64$ (Default) & $d = 128$   \\ \midrule
Complete $(z_{1:4})$ & Image $(z_{1})$ & $0.91 \pm 0.04$ & $0.96 \pm 0.02$ & $0.92 \pm 0.04$   \\
Complete $(z_{1:4})$ & Sound $(z_{2})$ & $0.77 \pm 0.17$ & $0.87 \pm 0.16$ & $0.96 \pm 0.04$ \\
Complete $(z_{1:4})$ & Trajectory $(z_{3})$ & $0.86 \pm 0.04$ & $0.86 \pm 0.05$ & $0.86 \pm 0.07$ \\
Complete $(z_{1:4})$ & Label $(z_{4})$ & $1.00 \pm 0.00$ & $1.00 \pm 0.00$ & $1.00 \pm 0.00$  \\\bottomrule
\end{tabular}
\label{table:ablation:intermediate:dca}
\vspace{-2ex}
\end{table}

\textbf{Temperature parameter} We study the performance of GMC when varying $\tau \in \{0.05, 0.1, 0.2, 0.3, 0.5\}$ (see Equation \eqref{eq:similarity}). We present the classification results and DCA scores in Table \ref{table:ablation:temperature:classification} and Table \ref{table:ablation:temperature:dca}, respectively. We observe that classification results are rather robust to different values of temperature, while increasing the temperature seems to have slightly negative effect on the geometry of the representations. For example, in Table \ref{table:ablation:temperature:dca}, we observe that for $\tau = 0.5$ the trajectory representations $z_3$ are worse aligned with $z_{1:4}$.

\textbf{Dimensionality of intermediate representations} We vary the dimension of the intermediate representations space ${d = \{32, 64, 128\}}$ and present the resulting classification results and DCA scores in Table \ref{table:ablation:intermediate:classification} and Table \ref{table:ablation:intermediate:dca}, respectively. The differences in classification results across different dimensions are covered by the margin of error, indicating the robustness of GMC to different sizes of the intermediate representations. We observe similar stability of the DCA scores in Table \ref{table:ablation:temperature:dca} with minor variations in the geometric alignment for the sound modality $z_2$ which benefits from the larger intermediate representation space. 

\begin{table}[!t]
\centering
\caption{Performance of GMC with different values of latent representation dimensionality $z \in \mathbb{R}^s$ in the MHD dataset, in a downstream classification task under complete and partial observations. Accuracy results averaged over 5 independent runs. Higher is better.}
\vspace{2ex}
\begin{tabular}{@{}lccc@{}}
\toprule
Observations & $d = 32$ & $d = 64$ (Default) & $d = 128$   \\ \midrule
Complete Observations & $\phantom{1}99.99 \pm 0.01$  & $100.00 \pm 0.00$  & $\phantom{1}99.99 \pm 0.01$  \\  \midrule
Image Observations & $\phantom{1}99.75 \pm 0.04$  & $\phantom{1}99.75 \pm 0.03$  & $\phantom{1}99.72 \pm 0.07$     \\
Sound Observations & $\phantom{1}93.31 \pm 0.41$  & $\phantom{1}93.04 \pm 0.45$  & $\phantom{1}93.34 \pm 0.51$   \\
Trajectory Observations & $\phantom{1}99.96 \pm 0.01$  & $\phantom{1}99.96 \pm 0.02$  & $\phantom{1}99.96 \pm 0.01$    \\
Label Observations & $100.00 \pm 0.00$  & $100.00 \pm 0.00$  & $100.00 \pm 0.00$  \\ \bottomrule
\end{tabular}
\label{table:ablation:latent:classification}
\vspace{-2ex}
\end{table}

\begin{table}[!t]
\centering
\caption{DCA score obtained on GMC representations when varying the dimension of latent representations $z \in \mathbb{R}^d$ in the MHD dataset, evaluating the geometric alignment of complete representations $z_{1:4}$ and modality-specific ones $\{z_1, \dots, z_4\}$ used as $R$ and $E$ inputs in DCA, respectively. The score is averaged over 5 independent runs. Higher is better.}
\vspace{2ex}
\begin{tabular}{@{}llccc@{}}
\toprule
 $R$ & $E$ & $d = 32$ & $d = 64$ (Default) & $d = 128$   \\ \midrule
Complete $(z_{1:4})$ & Image $(z_{1})$ & $0.93 \pm 0.03$ & $0.96 \pm 0.02$ & $0.91 \pm 0.03$   \\
Complete $(z_{1:4})$ & Sound $(z_{2})$ & $0.89 \pm 0.01$ & $0.87 \pm 0.16$ & $0.86 \pm 0.19$  \\
Complete $(z_{1:4})$ & Trajectory $(z_{3})$ & $0.81 \pm 0.03$ & $0.86 \pm 0.05$ & $0.88 \pm 0.06$ \\
Complete $(z_{1:4})$ & Label $(z_{4})$ & $1.00 \pm 0.00$ & $1.00 \pm 0.00$ & $1.00 \pm 0.00$  \\\bottomrule
\end{tabular}
\label{table:ablation:latent:dca}
\vspace{-2ex}
\end{table}

\textbf{Dimensionality of latent representations} We repeat a similar evaluation for the dimension of the latent space ${s = \{32, 64, 128\}}$ and present the classification and DCA scores in Table \ref{table:ablation:latent:classification} and Table \ref{table:ablation:latent:dca}, respectively. We observe that GMC is robust to changes in $s$ both in terms of performance and geometric alignment.

\begin{table}[!t]
\centering
\caption{Performance of GMC with different loss functions in the MHD dataset, in a downstream classification task under complete and partial observations. Accuracy results averaged over 5 independent runs. Higher is better.}
\vspace{2ex}
\begin{tabular}{@{}lcc@{}}
\toprule
Observations & $\mathcal{L}_{\text{GMC}}$ (Default) & $\mathcal{L}_{\text{GMC}}^{*}$  \\ \midrule
Complete Observations  & $100.00 \pm 0.00$ & $\phantom{1}99.97 \pm 0.02$    \\  \midrule
Image Observations &  $\phantom{1}99.75 \pm 0.03$ &  $\phantom{1}99.87 \pm 0.01$      \\
Sound Observations & $\phantom{1}93.04 \pm 0.45$ &  $\phantom{1}92.79 \pm 0.24$     \\
Trajectory Observations & $\phantom{1}99.96 \pm 0.02$ &  $\phantom{1}99.98 \pm 0.01$      \\
Label Observations & $100.00 \pm 0.00$  & $100.00 \pm 0.00$   \\ \bottomrule
\end{tabular}
\label{table:ablation:loss:classification}
\vspace{-2ex}
\end{table}

\begin{table}[!t]
\centering
\caption{DCA score obtained on GMC representations when trained different loss functions in the MHD dataset, evaluating the geometric alignment of complete representations $z_{1:4}$ and modality-specific ones $\{z_1, \dots, z_4\}$ used as $R$ and $E$ inputs in DCA, respectively. The score is averaged over 5 independent runs. Higher is better.}
\vspace{2ex}
\begin{tabular}{@{}llcc@{}}
\toprule
 $R$ & $E$ & $\mathcal{L}_{\text{GMC}}$ (Default) & $\mathcal{L}_{\text{GMC}}^{*}$   \\ \midrule
Complete $(z_{1:4})$ & Image $(z_{1})$ & $0.96 \pm 0.02$  & $0.80 \pm 0.02$  \\
Complete $(z_{1:4})$ & Sound $(z_{2})$ &  $0.87 \pm 0.16$  & $0.27 \pm 0.14$  \\
Complete $(z_{1:4})$ & Trajectory $(z_{3})$ & $0.86 \pm 0.05$  & $0.86 \pm 0.03$  \\
Complete $(z_{1:4})$ & Label $(z_{4})$  & $1.00 \pm 0.00$  & $0.24 \pm 0.10$  \\\bottomrule
\end{tabular}
\label{table:ablation:loss:dca}
\vspace{-2ex}
\end{table}

\textbf{Loss function} We consider an ablated version of the loss function, $\mathcal{L}_{\text{GMC}}^{*}$, that considers only complete-observations as negative pairs, i.e. $\Omega^{*}(i) = \sum_{i \neq j} s_{1:M, 1:M}(i, j)$ for $j = 1, \dots, B$ where $B$ is the size of the mini-batch. Due to the symmetry of negative pairs in this setting, we only consider positive pairs $(z_{m}^i, z_{1:M}^i)$. We present the classification results and DCA scores in Table \ref{table:ablation:loss:classification} and Table \ref{table:ablation:loss:dca}, respectively. The results in Table \ref{table:ablation:loss:classification} highlight the importance of the contrasting the complete representations to learn a robust representation suitable for downstream tasks as we observe minimal variation in classification accuracy when considering different loss. However, we observe worse geometric alignment when using $\mathcal{L}_{\text{GMC}}^{*}$ loss during training of GMC. This suggests that contrasting among individual modalities is beneficial for geometrical alignment of the representations.

\begin{figure*}[t]
    \centering
    \includegraphics[width=16cm]{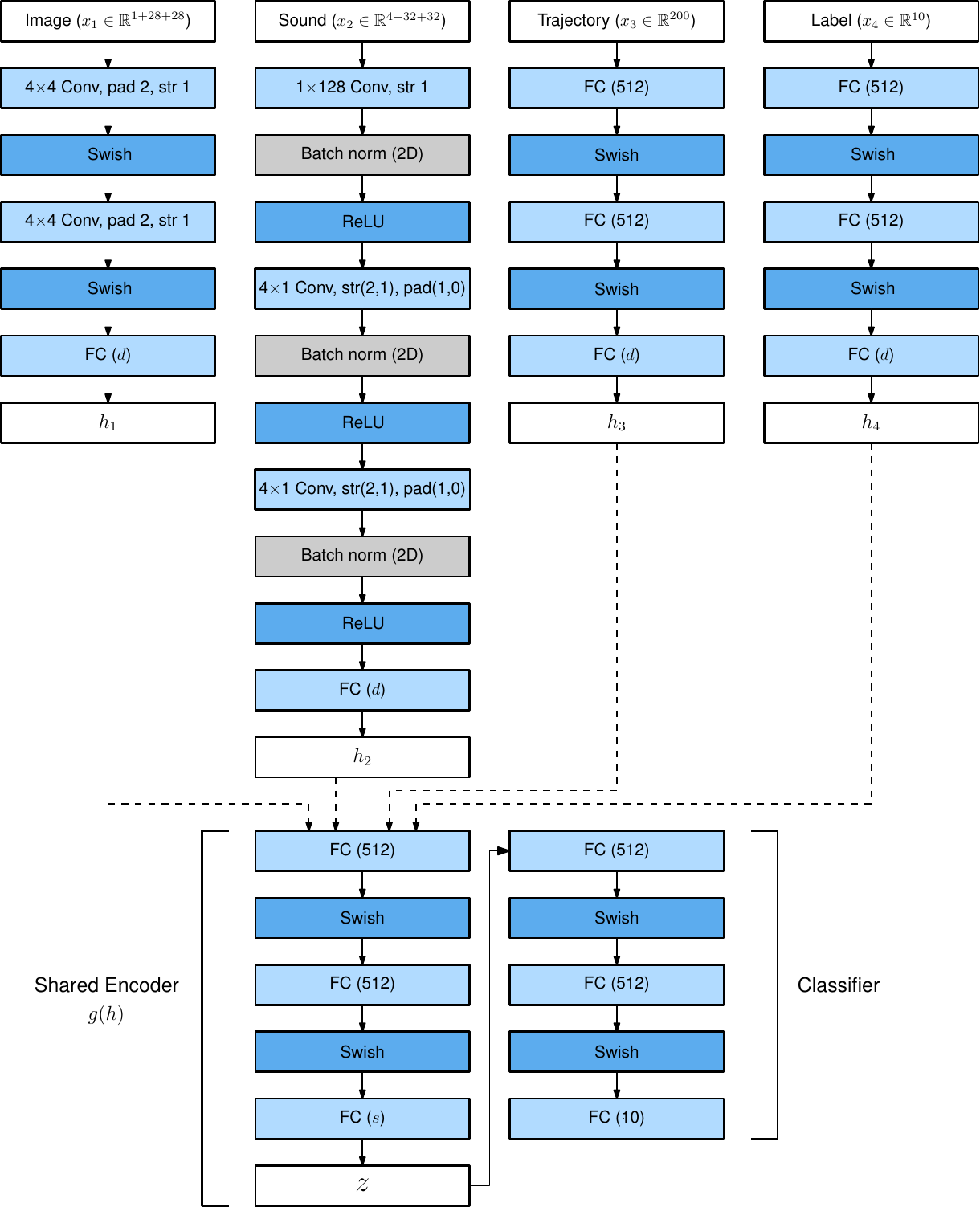}
    \caption{GMC model for the unsupervised experiment of Section \ref{sec:experiments:unsupervised}. Dashed lines represent potential connections between the intermediate representations $\{h_1, \ldots, h_4\}$ and the shared head $g(h)$. For the joint modality base encoder (not depicted) we employ an additional network with an identical architecture to the modality-specific ones, employing a late-fusion mechanism of all modalities before the projection (FC) to the intermediate representation $h$.}
    \label{fig:appendix:models:unsupervised}
\end{figure*}

\begin{figure*}[t]
    \centering
    \includegraphics[width=12cm]{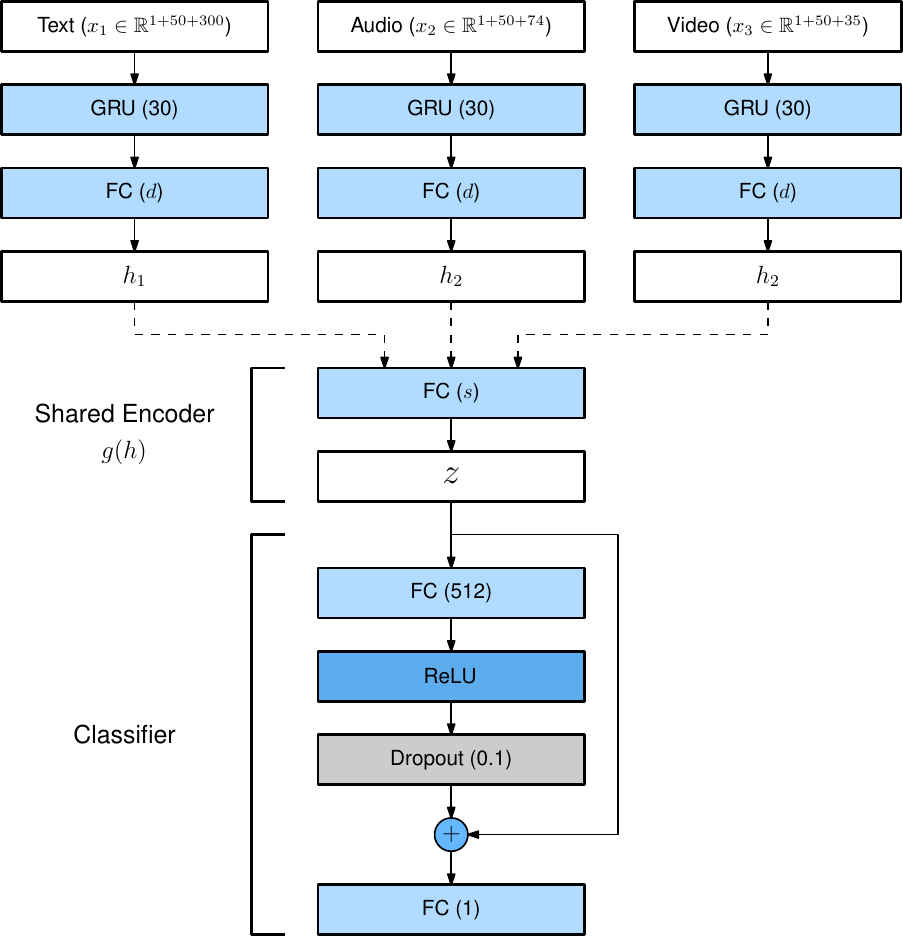}
    \caption{GMC model for the supervised experiment of Section \ref{sec:experiments:supervised}. Dashed lines represent potential connections between the intermediate representations $\{h_1, \ldots, h_3\}$ and the shared head $g(h)$. For the joint modality base encoder (not depicted) we employ the baseline multimodal transformer model, whose architecture we refer to \citet{tsai2019MULT}.}
    \label{fig:appendix:models:supervised}
\end{figure*}

\begin{figure*}[t]
    \centering
    \includegraphics[width=8cm]{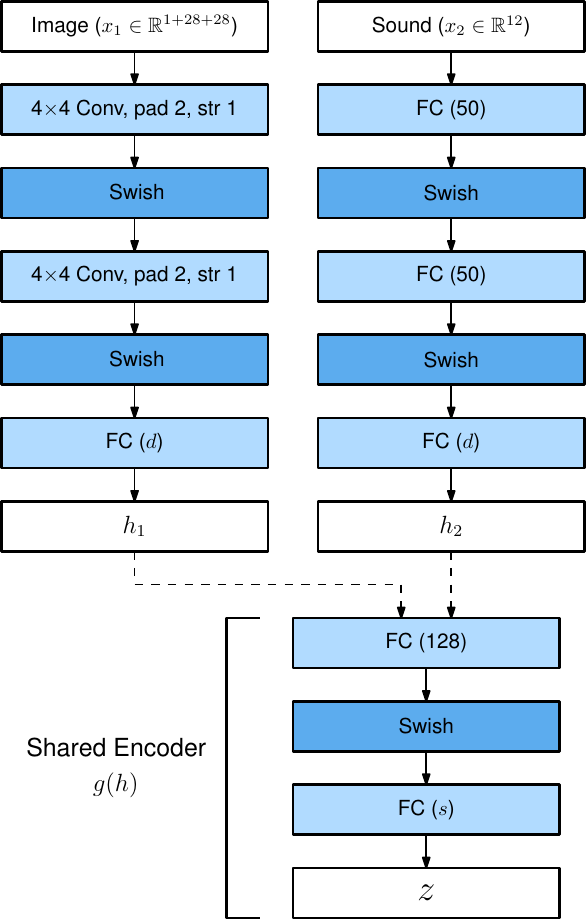}
    \caption{GMC model for the RL experiment of Section \ref{sec:experiments:rl}. Dashed lines represent potential connections between the intermediate representations $\{h_1, h_2\}$ and the shared head $g(h)$. For the joint modality base encoder (not depicted) we employ an additional network with an identical architecture to the modality-specific ones, employing a late-fusion mechanism of all modalities before the projection (FC) to the intermediate representation $h$. For the policy network, we refer to \citet{silva2019playing}.}
    \label{fig:appendix:models:rl}
\end{figure*}

\begin{table*}[t]
     \caption{Performance of different multimodal representation methods in the CMU-MOSI dataset, in a classification task under complete and partial observations. Results averaged over 5 independent runs. Arrows indicate the direction of improvement. }
     \centering
     \vspace{2ex} \hfill
    \begin{subtable}[h]{0.4\textwidth}
        \centering
        \begin{tabular}{@{}lcc@{}}
        \toprule
        Metric & Baseline & GMC (Ours) \\ \midrule
        MAE ($\downarrow$) & $1.033 \pm 0.037$   & $\bf 1.010 \pm 0.070$  \\
        Cor ($\uparrow$) & $0.642 \pm 0.008$  & $\bf 0.649 \pm 0.019$  \\
        F1  ($\uparrow$)& $0.770 \pm 0.017$  & $\bf 0.776 \pm 0.023$  \\
        Acc ($\%, \uparrow$)& $77.07 \pm 01.67$  & $\bf 77.59 \pm 02.20$  \\ \bottomrule
        \end{tabular}
       \caption{Complete Observations $(x_{1:3})$}
       \label{table:supervised:mosi:complete}
    \end{subtable}
    \hfill
    \begin{subtable}[h]{0.4\textwidth}
        \centering
        \begin{tabular}{@{}lcc@{}}
        \toprule
        Metric & Baseline & GMC (Ours) \\ \midrule
        MAE ($\downarrow$) & $1.244 \pm 0.100$   & $\bf 1.119 \pm 0.033$  \\
        Cor ($\uparrow$) & $0.431 \pm 0.208$  & $\bf 0.573 \pm 0.016$  \\
        F1  ($\uparrow$)& $0.698 \pm 0.053$  & $\bf 0.727 \pm 0.013$  \\
        Acc ($\%, \uparrow$)& $66.28 \pm 07.74$  & $\bf 72.32 \pm 0.013$  \\ \bottomrule
        \end{tabular}
        \caption{Text Observations $(x_1)$}
        \label{table:supervised:mosi:text}
     \end{subtable}
     \hfill
      \vspace{1ex}

     \hfill
     \begin{subtable}[h]{0.4\textwidth}
        \centering
         \begin{tabular}{@{}lcc@{}}
        \toprule
        Metric & Baseline & GMC (Ours) \\ \midrule
        MAE ($\downarrow$) & $1\bf .431 \pm 0.025$   & $1.434 \pm 0.017$  \\
        Cor ($\uparrow$) & $0.056 \pm 0.071$  & $\bf 0.211 \pm 0.010$  \\
        F1  ($\uparrow$)& $\bf 0.588 \pm 0.076$  & $0.570 \pm 0.006$  \\
        Acc ($\%, \uparrow$)& $47.20 \pm 05.67$  & $\bf 55.91 \pm 01.11$  \\ \bottomrule
        \end{tabular}
        \caption{Audio Observations $(x_2)$}
        \label{table:supervised:mosi:audio}
     \end{subtable}
     \hfill
    \begin{subtable}[h]{0.4\textwidth}
        \centering
         \begin{tabular}{@{}lcc@{}}
        \toprule
        Metric & Baseline & GMC (Ours) \\ \midrule
        MAE ($\downarrow$) & $\bf 1.406 \pm 0.041$   & $1.452 \pm 0.035$  \\
        Cor ($\uparrow$) & $0.021 \pm 0.028$  & $\bf 0.176 \pm 0.028$  \\
        F1  ($\uparrow$)& $\bf 0.659 \pm 0.049$  & $0.550 \pm 0.015$  \\
        Acc ($\%, \uparrow$)& $53.87 \pm 05.77$  & $\bf 54.30 \pm 01.96$  \\ \bottomrule
        \end{tabular}
        \caption{Video Observations $(x_3)$}
        \label{table:supervised:mosi:video}
     \end{subtable}
     \hfill
     \label{table:supervised:mosi}
\end{table*}

\begin{table}[t]
\centering
\caption{DCA score of the models in the CMU-MOSI dataset, evaluating the geometric alignment of complete representations $z_{1:4}$ and modality-specific ones $\{z_1, z_2, z_3\}$ used as $R$ and $E$ inputs in DCA, respectively. The score is averaged over 5 independent runs. Higher is better.}
\vspace{1ex}
\begin{tabular}{@{}lccc@{}}
\toprule
\multicolumn{1}{c}{R} & \multicolumn{1}{c}{E} & \multicolumn{1}{c}{Baseline} & \multicolumn{1}{c}{\textbf{GMC (Ours)}} \\ \midrule
Complete $(z_{1:3})$ & Text $(z_{1})$ & $0.54 \pm 0.07$ & $\bf 0.93 \pm 0.02$  \\
Complete $(z_{1:3})$ & Audio $(z_{2})$ & $0.14 \pm 0.06$ & $\bf 0.75 \pm 0.05$ \\
Complete $(z_{1:3})$ & Vision $(z_{3})$ & $0.36 \pm 0.09$ & $\bf 0.85 \pm 0.04$  \\\bottomrule
\end{tabular}
\label{table:mosi:dca}
\end{table}

\section{Experiment 2: Supervised Learning with the CMU-MOSI dataset}
\label{sec:appendix:mosi} 

In this section, we repeat the experimental evaluation of Section \ref{sec:experiments:supervised} with the CMU-MOSI dataset. We employ the same baseline and GMC architectures as in the CMU-MOSEI evaluation and consider the same evaluation setup.

\textbf{Results} The results obtained on CMU-MOSI are reported in Table~\ref{table:supervised:mosi}. We observe that GMC improves the robustness of the model to the missing modalities as seen from Tables~\ref{table:supervised:mosi:text},~\ref{table:supervised:mosi:audio} and~\ref{table:supervised:mosi:video} where we use only individual modalities as inputs. However, the increase in performance is not as signification as in the case of the CMU-MOSEI dataset for audio ($x_2$) and video ($x_3$) modalities where the baseline outperforms GMC on MAE and F1 scores. We hypothesise that this behaviour is due to the intrinsic difficulty of forming good contrastive pairs in small-sizes datasets \cite{cao2021rethinking}: the CMU-MOSI dataset has only 1513 training samples which hinders the learning of a quality latent representations.
However, we observe that GMC still significantly improves the geometric alignment (Table \ref{table:mosi:dca}) of the modality-specific representations $z_m$ (comprising the set $E$) and complete representations $z_{1:3}$ (comprising the set $R$) compared to the baseline, even in this regime of small data.

\section{Model Architecture}
\label{sec:appendix:architecture}
We report the model architectures for GMC employed in our work: in Figure~\ref{fig:appendix:models:unsupervised} we present the model employed for the unsupervised experiment of Section \ref{sec:experiments:unsupervised}; in Figure~\ref{fig:appendix:models:supervised} we present the model employed for the supervised experiment of Section \ref{sec:experiments:supervised}; in Figure~\ref{fig:appendix:models:rl} we present the model employed in the RL experiment of Section \ref{sec:experiments:rl}.

\clearpage
\newpage
\newpage
\newpage
\section{Training Hyperparameters}
\label{sec:appendix:hyperparameters}

In Table \ref{table:hyperparameters} we present the hyperparameters employed in this work. For training the controller in the RL task, we employ the same training hyperparameters as in \citet{silva2019playing}.

\section{Additional Visualizations of the Alignment of Complete and Modality-Specific Representations}
\label{sec:appendix:vis}

\begin{figure*}[t]
    \centering
    \begin{subfigure}[b]{0.31\textwidth}
        \centering
        \includegraphics[height=2.8cm]{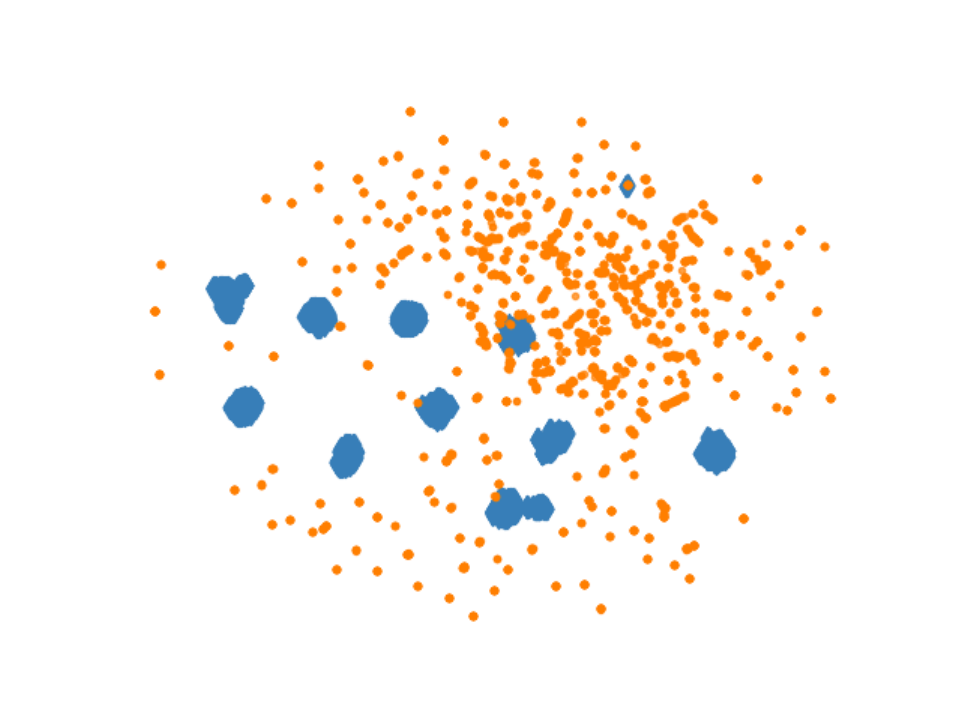}
        \caption{MVAE}
        \label{fig:appendix:alignment:sound:mvae}
    \end{subfigure}
    \hfill
    \begin{subfigure}[b]{0.31\textwidth}
        \centering
        \includegraphics[height=2.8cm]{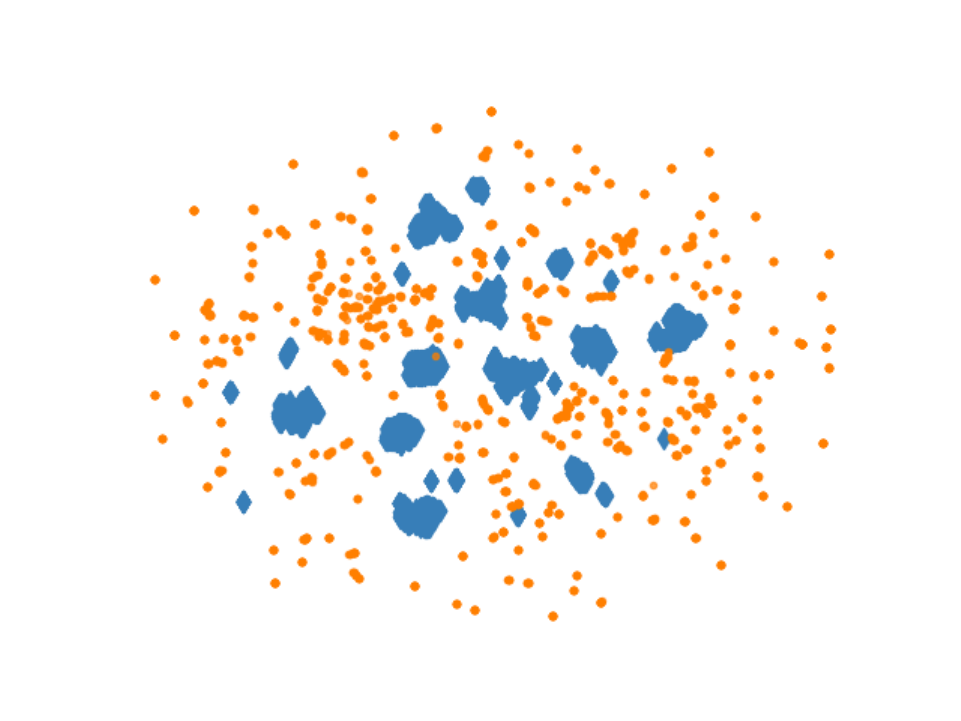}
        \caption{MMVAE}
        \label{fig:appendix:alignment:sound:mmvae}
    \end{subfigure}
    \hfill
    \begin{subfigure}[b]{0.31\textwidth}
        \centering
        \includegraphics[height=2.8cm]{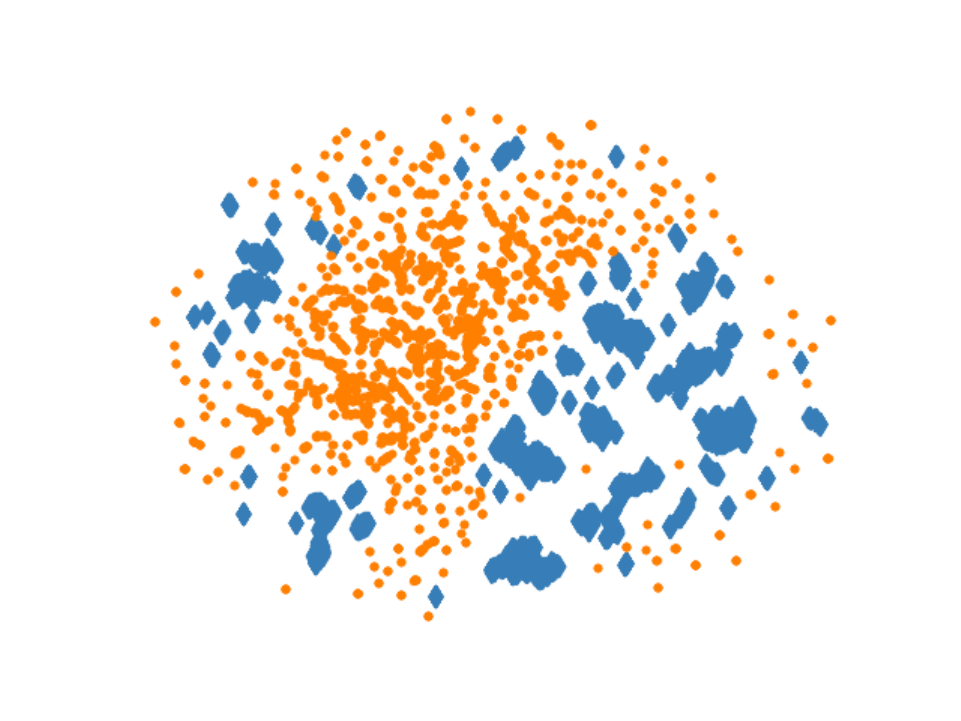}
        \caption{Nexus}
        \label{fig:appendix:alignment:sound:nexus}
    \end{subfigure}
    
    \vspace{1ex}
    
    \begin{subfigure}[b]{0.31\textwidth}
        \centering
        \includegraphics[height=2.8cm]{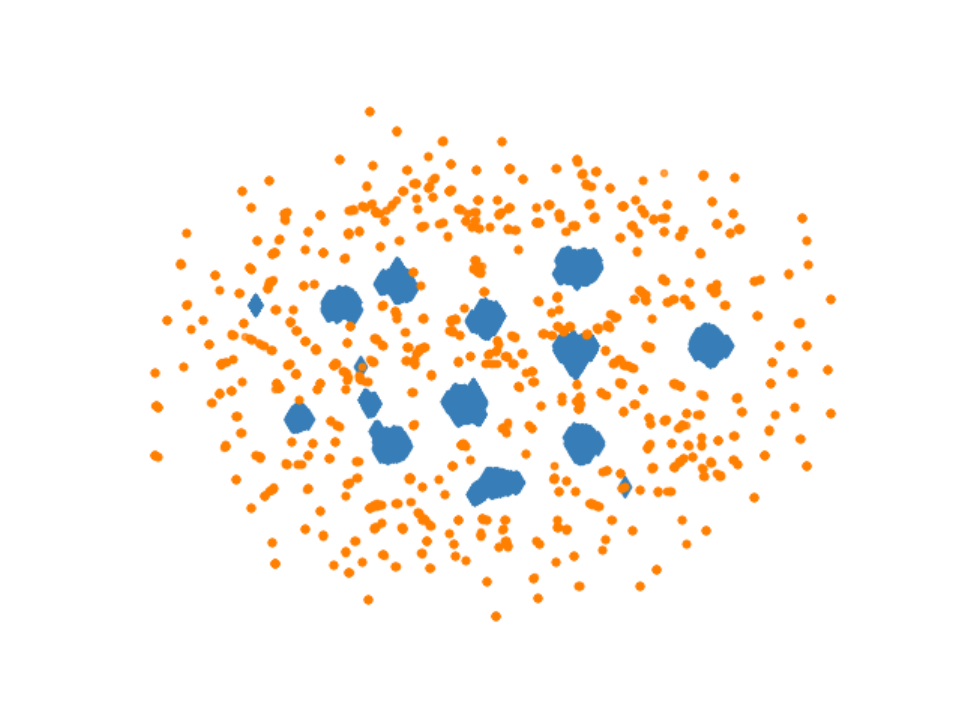}
        \caption{MUSE}
        \label{fig:appendix:alignment:sound:muse}
    \end{subfigure}
    \hfill
    \begin{subfigure}[b]{0.31\textwidth}
        \centering
        \includegraphics[height=2.8cm]{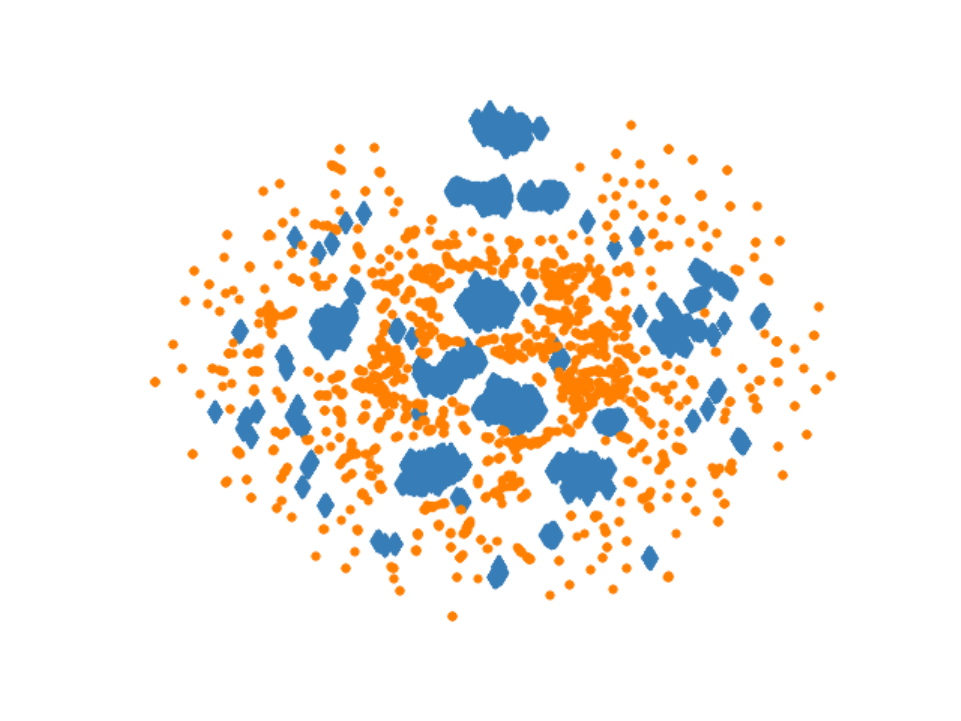}
        \caption{MFM}
        \label{fig:appendix:alignment:sound:mfm}
    \end{subfigure}
    \hfill
    \begin{subfigure}[b]{0.31\textwidth}
        \centering
        \includegraphics[height=2.8cm]{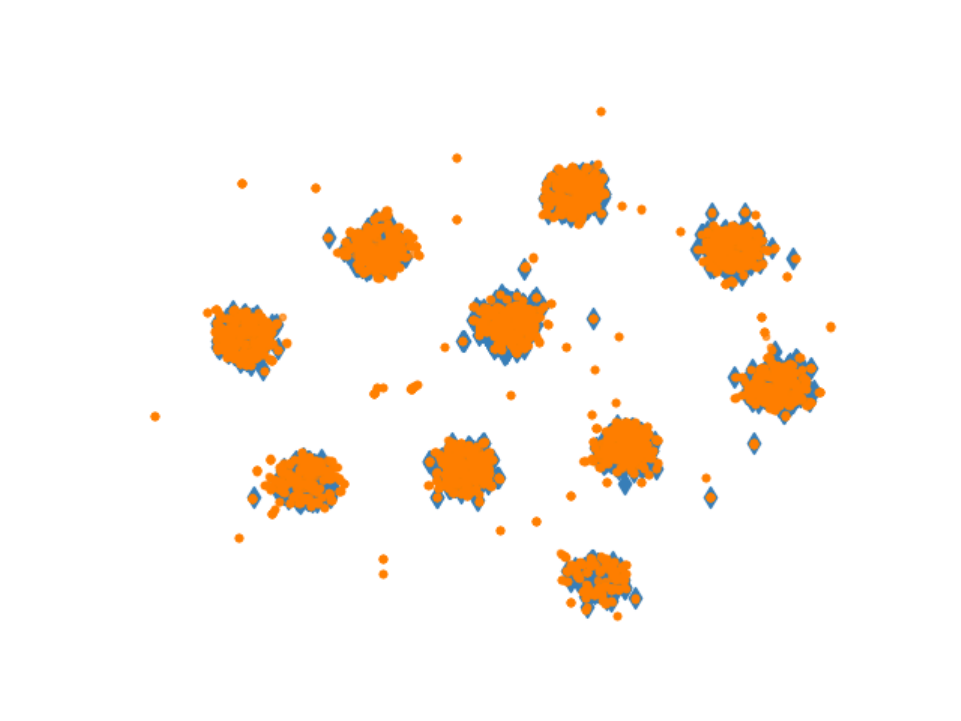}
        \caption{\textbf{GMC (Ours)}}
        \label{fig:appendix:alignment:sound:gmc}
    \end{subfigure}
    \caption{UMAP visualization of complete representations $z_{1:4}$ (blue) and sound representations $z_2$ (orange) obtained from several state-of-the-art multimodal representation learning models on the MHD dataset considered in Section~\ref{sec:experiments:unsupervised}. Best viewed in color.}
    \label{fig:appendix:alignment:sound}
\end{figure*}

\begin{figure*}[t]
    \centering
    \begin{subfigure}[b]{0.31\textwidth}
        \centering
        \includegraphics[height=2.8cm]{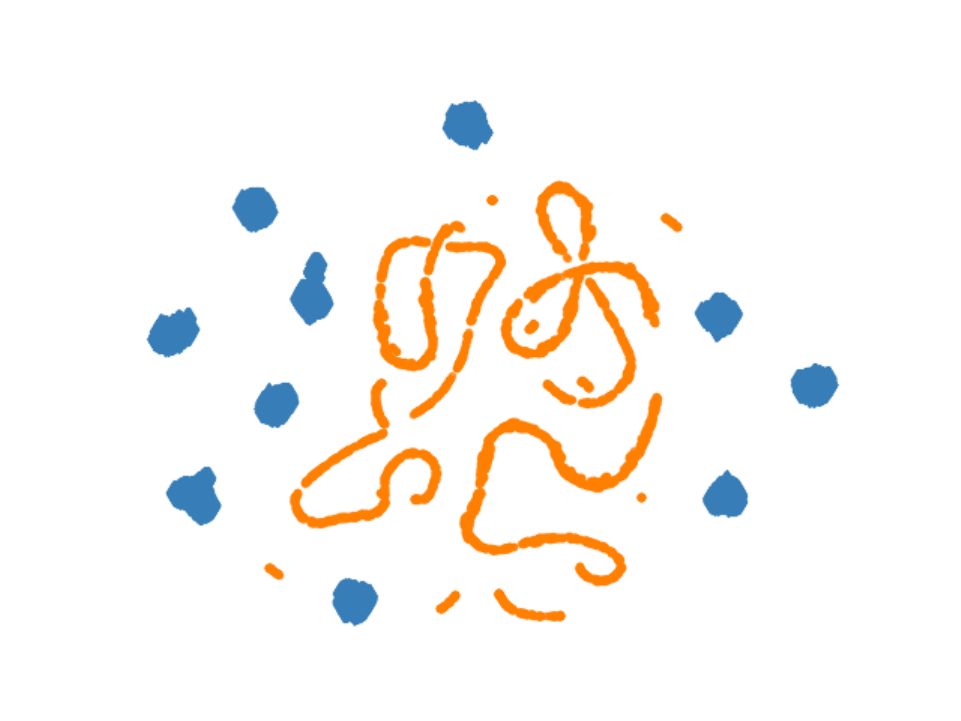}
        \caption{MVAE}
        \label{fig:appendix:alignment:motion:mvae}
    \end{subfigure}
    \hfill
    \begin{subfigure}[b]{0.31\textwidth}
        \centering
        \includegraphics[height=2.8cm]{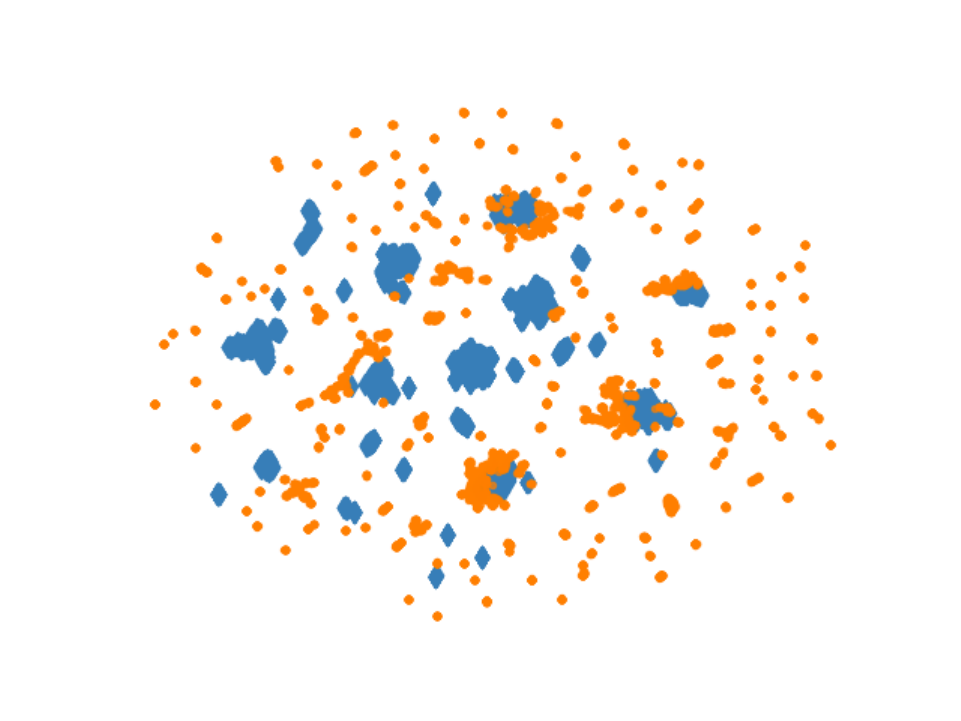}
        \caption{MMVAE}
        \label{fig:appendix:alignment:motion:mmvae}
    \end{subfigure}
    \hfill
    \begin{subfigure}[b]{0.31\textwidth}
        \centering
        \includegraphics[height=2.8cm]{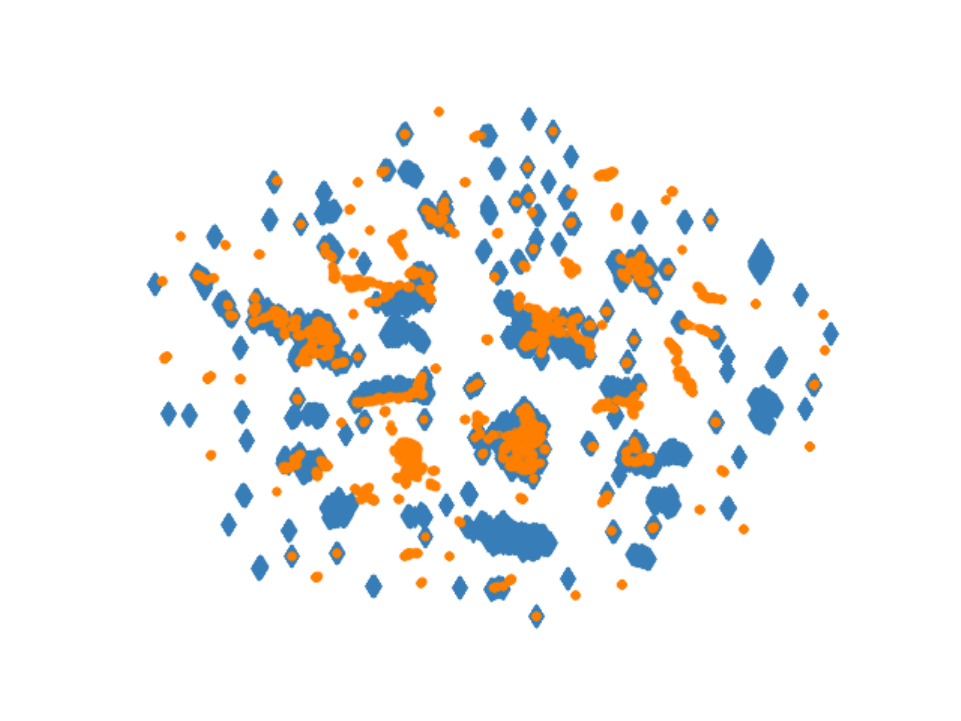}
        \caption{Nexus}
        \label{fig:appendix:alignment:motion:nexus}
    \end{subfigure}
    
    \vspace{1ex}
    
    \begin{subfigure}[b]{0.31\textwidth}
        \centering
        \includegraphics[height=2.8cm]{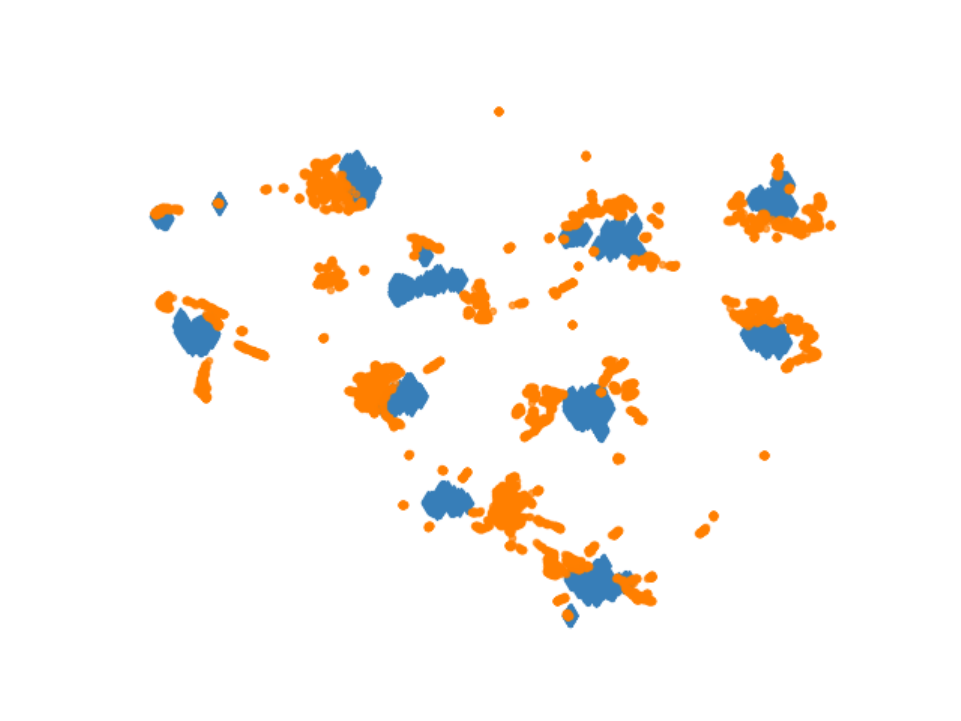}
        \caption{MUSE}
        \label{fig:appendix:alignment:motion:muse}
    \end{subfigure}
    \hfill
    \begin{subfigure}[b]{0.31\textwidth}
        \centering
        \includegraphics[height=2.8cm]{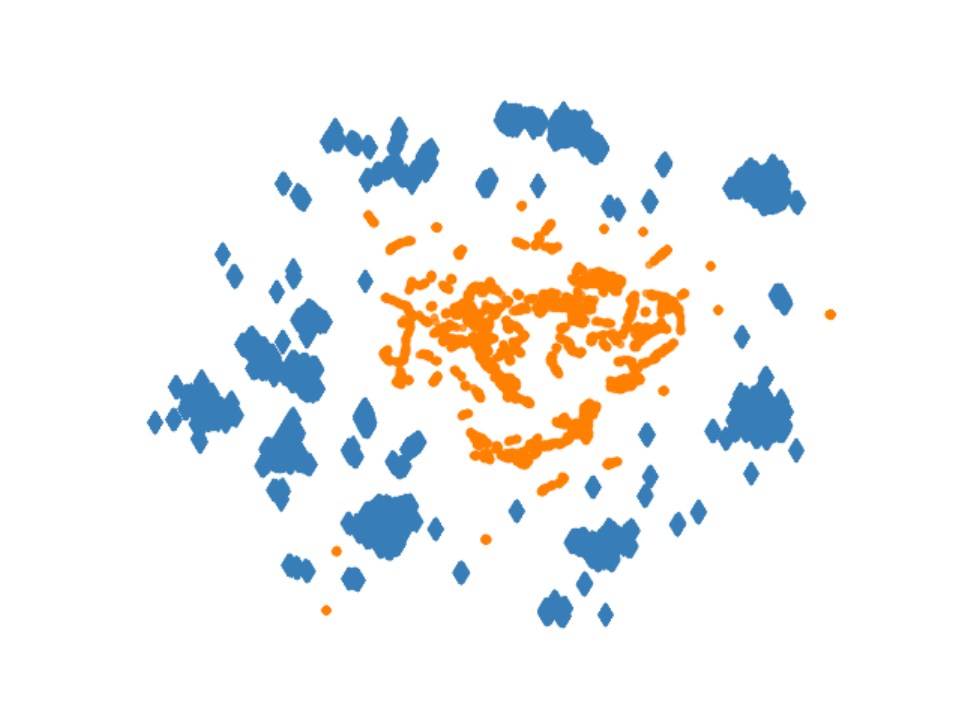}
        \caption{MFM}
        \label{fig:appendix:alignment:motion:mfm}
    \end{subfigure}
    \hfill
    \begin{subfigure}[b]{0.31\textwidth}
        \centering
        \includegraphics[height=2.8cm]{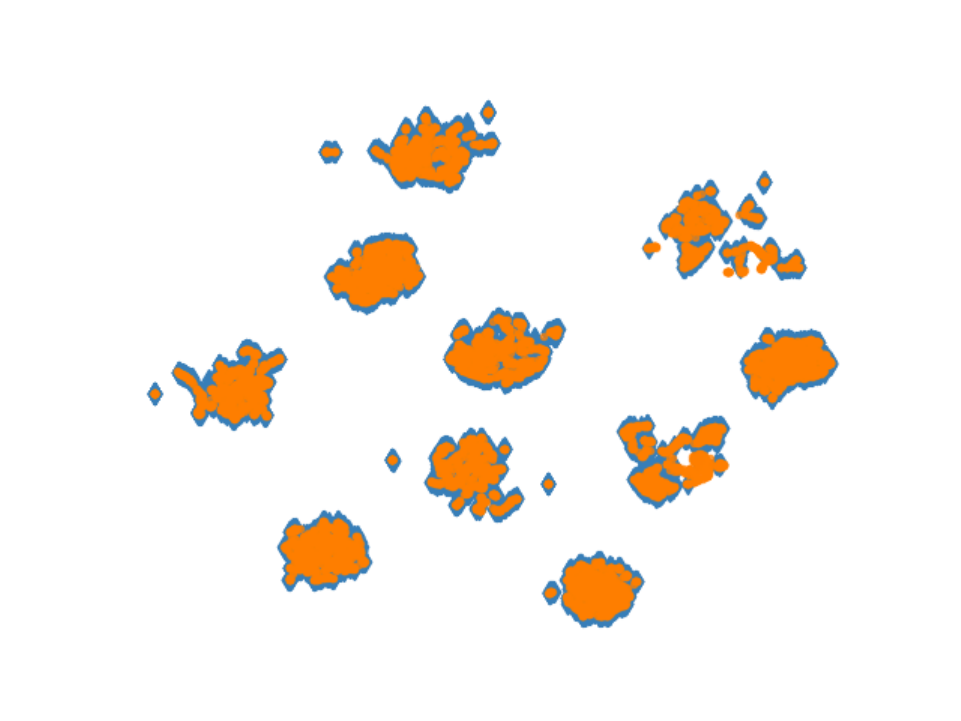}
        \caption{\textbf{GMC (Ours)}}
        \label{fig:appendix:alignment:motion:gmc}
    \end{subfigure}
    \caption{UMAP visualization of complete representations $z_{1:4}$ (blue) and trajectory representations $z_3$ (orange) obtained from several state-of-the-art multimodal representation learning models on the MHD dataset considered in Section~\ref{sec:experiments:unsupervised}. Best viewed in color.}
    \label{fig:appendix:alignment:motion}
\end{figure*}

\begin{figure*}[t]
    \centering
    \begin{subfigure}[b]{0.31\textwidth}
        \centering
        \includegraphics[height=2.8cm]{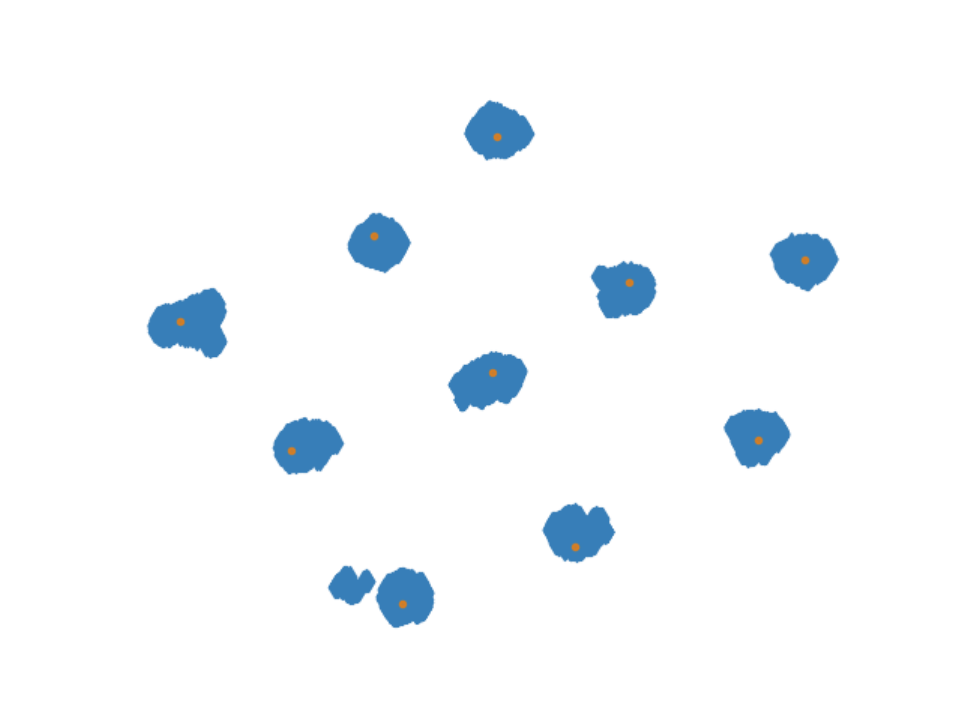}
        \caption{MVAE}
        \label{fig:appendix:alignment:label:mvae}
    \end{subfigure}
    \hfill
    \begin{subfigure}[b]{0.31\textwidth}
        \centering
        \includegraphics[height=2.8cm]{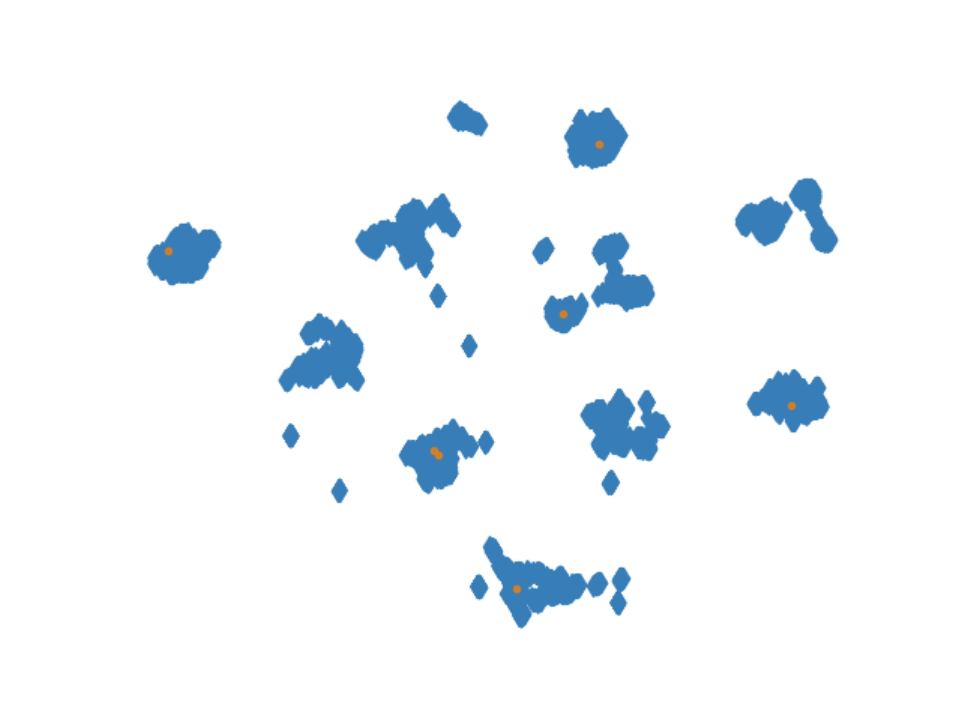}
        \caption{MMVAE}
        \label{fig:appendix:alignment:label:mmvae}
    \end{subfigure}
    \hfill
    \begin{subfigure}[b]{0.31\textwidth}
        \centering
        \includegraphics[height=2.8cm]{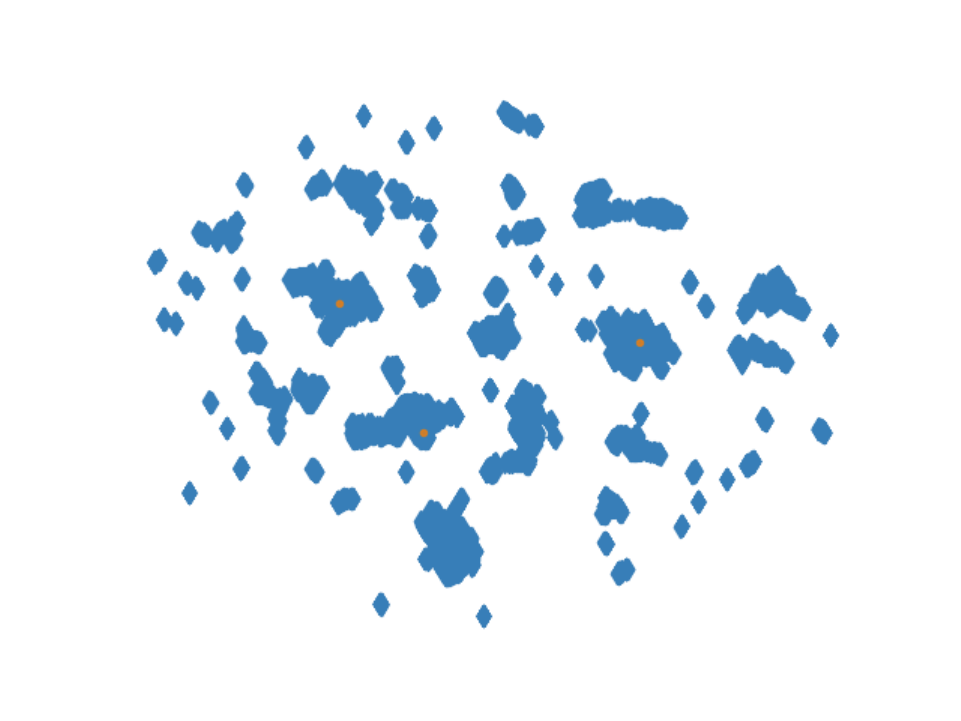}
        \caption{Nexus}
        \label{fig:appendix:alignment:label:nexus}
    \end{subfigure}
    
    \vspace{1ex}
    
    \begin{subfigure}[b]{0.31\textwidth}
        \centering
        \includegraphics[height=2.8cm]{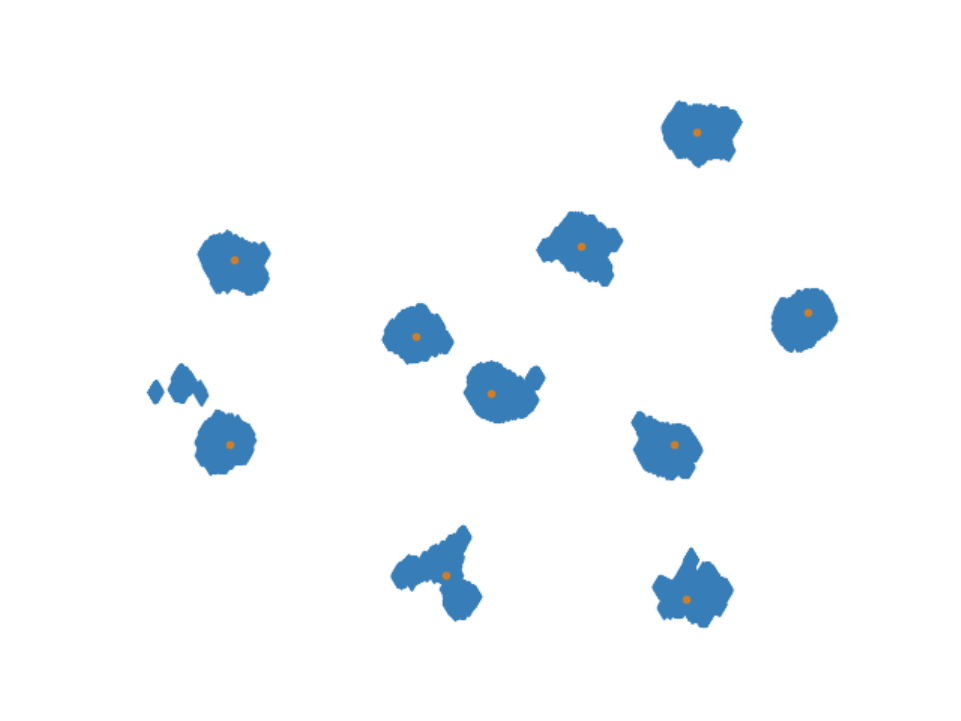}
        \caption{MUSE}
        \label{fig:appendix:alignment:label:muse}
    \end{subfigure}
    \hfill
    \begin{subfigure}[b]{0.31\textwidth}
        \centering
        \includegraphics[height=2.8cm]{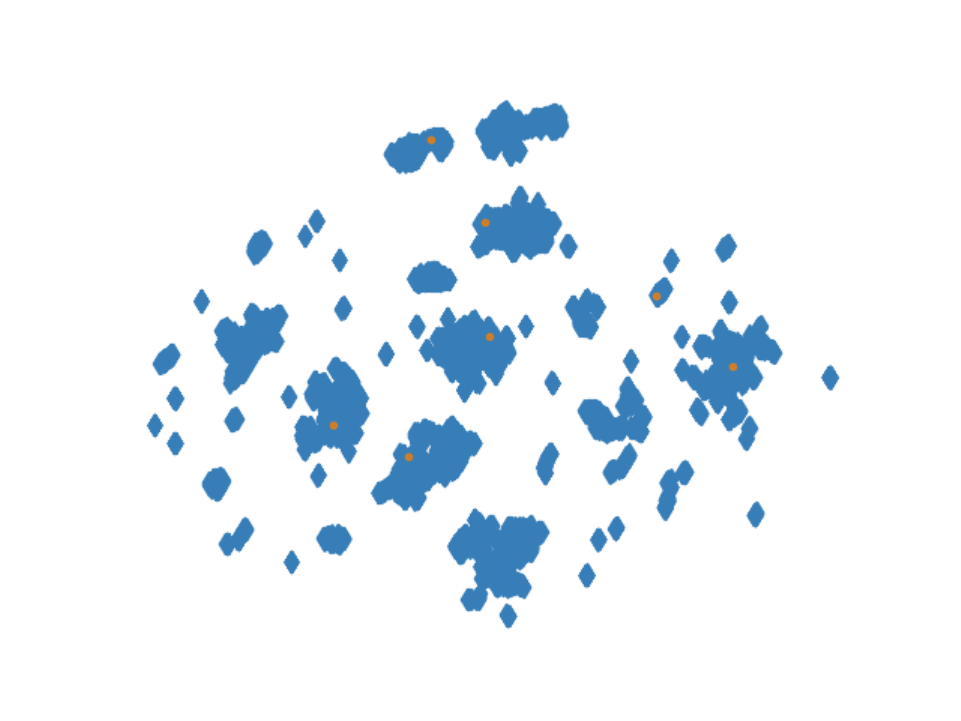}
        \caption{MFM}
        \label{fig:appendix:alignment:label:mfm}
    \end{subfigure}
    \hfill
    \begin{subfigure}[b]{0.31\textwidth}
        \centering
        \includegraphics[height=2.8cm]{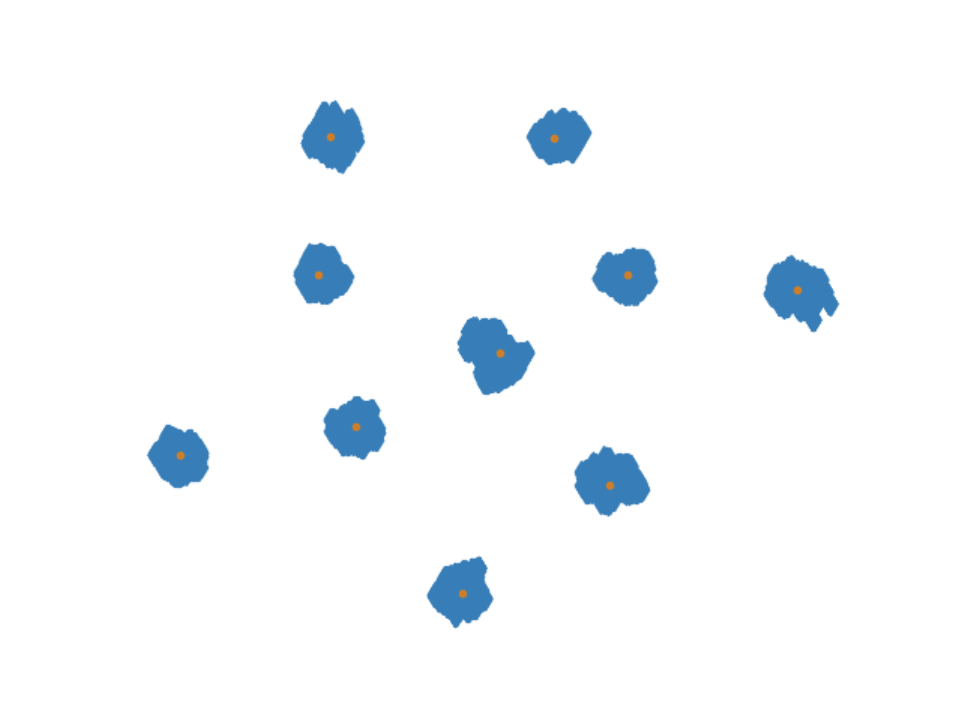}
        \caption{\textbf{GMC (Ours)}}
        \label{fig:appendix:alignment:label:gmc}
    \end{subfigure}
    \caption{UMAP visualization of complete representations $z_{1:4}$ (blue) and label representations $z_4$ (orange) obtained from several state-of-the-art multimodal representation learning models on the MHD dataset considered in Section~\ref{sec:experiments:unsupervised}. Best viewed in color.}
    \label{fig:appendix:alignment:label}
\end{figure*}

We present additional visualizations of encodings of complete and modality-specific representations in the MHD dataset for multiple multimodal representation models. In Figures~\ref{fig:appendix:alignment:sound}, \ref{fig:appendix:alignment:motion} and \ref{fig:appendix:alignment:label}, we show visualizations of sound representations $z_2$, trajectory $z_3$ and label $z_4$ (in orange), respectively, and complete representations $z_{1:4}$ (in blue). Note that points detected as outliers by DCA are \emph{not} included in the visualization. For example, we observe that certain labels representations for baseline models are marked as outliers in Figure~\ref{fig:appendix:alignment:label}.

\begin{table*}[b]
\centering
\caption{Training hyperparameters of GMC.}
\vspace{2ex}
\begin{subtable}{0.26\textwidth}
\centering
\caption{Unsupervised (Section~\ref{sec:experiments:unsupervised})}
\begin{tabular}{@{}lc@{}}
\toprule
Parameter & Value \\ \midrule
Intermediate size $d$ & 64 \\
Latent size $s$ & 64 \\
Model training epochs & 100 \\
Classifier training epochs & 50 \\
Learning rate & 1$\mathrm{e}{-3}$ \\
Batch size $B$ & 64 \\
Temperature $\tau$ & 0.1 \\ \bottomrule
\end{tabular}
\end{subtable}
\hfill
\begin{subtable}{0.31\textwidth}
\centering
\caption{Supervised (Section~\ref{sec:experiments:supervised})}
\begin{tabular}{@{}lc@{}}
\toprule
Parameter & Value \\ \midrule
Intermediate size $d$ & 60 \\
Latent size $s$ & 60 \\
Model training epochs & 40 \\
Learning rate & 1$\mathrm{e}{-3}$ (Decay) \\
Batch size $B$ & 40 \\
Temperature $\tau$ & 0.3 \\ \bottomrule
\end{tabular}
\vspace{2.5ex}
\end{subtable}
\hfill
\begin{subtable}{0.31\textwidth}
\centering
\caption{RL (Section~\ref{sec:experiments:rl})}
\begin{tabular}{@{}lc@{}}
\toprule
Parameter & Value \\ \midrule
Intermediate size $d$ & 64 \\
Latent size $s$ & 10 \\
Model training epochs & 500 \\
Learning rate & 1$\mathrm{e}{-3}$ (Decay) \\
Batch size $B$ & 128 \\
Temperature $\tau$ & 0.3 \\ \bottomrule
\end{tabular}
\vspace{2.5ex}
\end{subtable}
\label{table:hyperparameters}
\end{table*}

\end{document}